\documentclass{article}

\usepackage[preprint]{style}

\usepackage[utf8]{inputenc} 
\usepackage[T1]{fontenc}    
\usepackage{url}            
\usepackage{booktabs}       
\usepackage{amsfonts}       
\usepackage{nicefrac}       
\usepackage{microtype}      
\usepackage[table]{xcolor}

\title{Using Synthetic Images to Augment Small Medical Image Datasets}

\author{%
  Minh H. Vu \\
  Department of Diagnostics and Intervention \\
  Umeå University \\
  \texttt{minh.vu@umu.se} \\
  \And
  Lorenzo Tronchin \\
  Università Campus Bio-Medico di Roma \\
  \texttt{l.tronchin@unicampus.it} \\
  \And
  Tufve Nyholm \\
  Department of Diagnostics and Intervention \\
  Umeå University \\
  \texttt{tufve.nyholm@umu.se} \\
  \And
  Tommy Löfstedt \\
  Department of Computing Science \\
  Umeå University \\
  \texttt{tommy.lofstedt@umu.se} \\
}

\usepackage{lipsum}

\usepackage{amsmath}
\usepackage{amssymb}
\usepackage{bbm}
\usepackage{eucal}
\usepackage{enumitem}
\usepackage{booktabs}
\usepackage[labelfont=bf]{caption,subcaption}
\usepackage{bbm}
\usepackage{tikz}
\usepackage{enumitem}
\usepackage{lscape}
\usepackage{adjustbox}
\usepackage{graphicx}
\usepackage{makecell}
\usepackage{pifont}
\usepackage{float}
\usepackage{xspace}

\usepackage{tcolorbox}
\usepackage{nicefrac,xfrac}

\usepackage{array}
\usepackage{multirow}
\newcolumntype{L}[1]{>{\raggedright\let\newline\\\arraybackslash\hspace{0pt}}m{#1}}
\newcolumntype{C}[1]{>{\centering\let\newline\\\arraybackslash\hspace{0pt}}m{#1}}
\newcolumntype{R}[1]{>{\raggedleft\let\newline\\\arraybackslash\hspace{0pt}}m{#1}}

\newcommand{\mycomment}[1]{}

\definecolor{light-gray}{gray}{0.7}
\newcommand{\cmark}{\text{\ding{51}}}%
\newcommand{\xmark}{\color{light-gray} \text{\ding{55}}}%
\usepackage[acronym]{glossaries}

\newacronym[plural=FCNs,firstplural=Fully Convolutional Networks (FCNs)]{fcn}{FCN}{Fully Convolutional Network}
\newacronym{ml}{ML}{machine learning}
\newacronym{dl}{DL}{deep learning}
\newacronym{drs}{DRS}{Diabetic Retinopathy Screening}
\newacronym{vqa}{VQA}{Visual Question Answering}
\newacronym{slam}{SLAM}{Simultaneous Localization and Mapping}

\newacronym[plural=DNNs,firstplural=deep neural networks (DNNs)]{dnn}{DNN}{deep neural network}
\newacronym[plural=CNNs,firstplural=Convolutional Neural Networks (CNNs)]{cnn}{CNN}{Convolutional Neural Network}

\newacronym{cv}{CV}{computer vision}
\newacronym{nlp}{NLP}{natural language processing}
\newacronym{mcb}{MCB}{Multimodal Compact Bilinear}
\newacronym{mlb}{MLB}{Multimodal Low-rank Bilinear}
\newacronym{mutan}{MUTAN}{Multimodal Tucker Fusion for Visual Question Answering}
\newacronym{idrid}{IDRID}{Indian Diabetic Retinopathy Image Dataset}
\newacronym[plural=RNNs,firstplural=Recurrent Neural Networks (RNNs)]{rnn}{RNN}{Recurrent Neural Network}
\newacronym{lstm}{LSTM}{Long Short-term Memory}
\newacronym{bow}{BOW}{bag-of-words}
\newacronym{gru}{GRU}{Gated Recurrent Units}
\newacronym[plural=QAs,firstplural=Questions \& Answers (QAs)]{qa}{QA}{Question \& Answer}
\newacronym{ma}{MA}{Microaneurysms}
\newacronym{he}{HE}{Hemorrhages}
\newacronym{ex}{EX}{Hard Exudates}
\newacronym{se}{SE}{Soft Exudates}
\newacronym{nar}{NAR}{no apparent retinopathy}
\newacronym{dr}{DR}{Diabetic Retinopathy}
\newacronym{dme}{DME}{Diabetic Macular Edema}
\newacronym{pdr}{PDR}{Proliferative Diabetic Retinopathy}
\newacronym{wmlb}{WMLB}{Weighted Multimodal Low-rank Bilinear Attention Network}

\newacronym{qcmlb}{QC-MLB}{Question-Centric Multimodal Low-rank Bilinear}
\newacronym{bert}{BERT}{Bidirectional Encoder Representations from Transformers}
\newacronym{bleu}{BLEU}{Bilingual Evaluation Understudy}
\newacronym{mlm}{MLM}{Masked Language Model}
\newacronym{nsp}{NSP}{Next Sentence Prediction}
\newacronym{relu}{\textsc{ReLU}}{rectified linear unit}
\newacronym{nn}{NN}{neural network}
\newacronym{chal}{ImageCLEF-VQA-Med}{ImageCLEF-VQA-Med}
\newacronym{proposed}{\textit{\textless~Model~\textgreater}}{\textbf{Full name of the proposed model}}
\newacronym{mri}{MRI}{magnetic resonance imaging}

\newacronym{brats}{BraTS20}{Brain Tumors in Multimodal Magnetic Resonance Imaging Challenge 2020}
\newacronym{kits}{KiTS19}{Kidney Tumor Segmentation Challenge 2019}
\newacronym{ibsr}{IBSR18}{Internet Brain Segmentation Repository}
\newacronym{hene}{U-HAND}{Ume{\aa} Head and Neck Database}
\newacronym{pros}{U-PRO}{Ume{\aa} Pelvic Region Organs}
\newacronym{heart}{D-HEART}{Heart Segmentation Decathlon}
\newacronym{spleen}{D-SPLEEN}{Spleen Segmentation Decathlon}
\newacronym{lung}{D-LUNG}{Lung Segmentation Decathlon}
\newacronym{hippo}{D-HIPPO}{Hippocampus Segmentation Decathlon}
\newacronym{cifar100}{CIFAR-100}{Canadian Institute for Advanced Research, 100 classes}
\newacronym{mnist}{MNIST}{Modified National Institute of Standards and Technology database}

\newacronym{hpc2n}{HPC2N}{High Performance Computer Center North}
\newacronym{snic}{SNIC}{Swedish National Infrastructure for Computing}
\newacronym{flops}{FLOPs}{floating point operations}

\newacronym{ct}{CT}{computed tomography}
\newacronym{t1c}{T1c}{post-contrast T1-weighted}
\newacronym{t2}{T2w}{T2-weighted}
\newacronym{t1}{T1w}{T1-weighted}
\newacronym{flair}{FLAIR}{T2  Fluid  Attenuated  Inversion  Recovery}
\newacronym{lgg}{LGG}{low grade glioma}
\newacronym{hgg}{HGG}{high grade glioma}

\newacronym{dsc}{DSC}{S{\o}rensen-Dice coefficient}
\newacronym{seb}{SEB}{Squeeze-and-Excitation block}
\newacronym[plural=REBs,firstplural=ResNet blocks]{res}{REB}{ResNet block}
\newacronym{dauc}{DAUC}{Dice Area Under Curve}
\newacronym{rftp}{RFTPs}{Ratio of Filtered True Positives}
\newacronym[plural=SDs,firstplural=standard deviations (SDs)]{sd}{SD}{standard deviation}
\newacronym{ce}{CE}{categorical cross--entropy}
\newacronym[plural=GPUs,firstplural=Graphical Processing Units (GPUs)]{gpu}{GPU}{Graphical Processing Unit}
\newacronym[plural=HDs]{hd}{HD}{Hausdorff distance}
\newacronym{hd95}{HD95}{$95$-th percentile of the Hausdorff distance}
\newacronym{ravd}{RAVD}{relative absolute volume difference}
\newacronym{assd}{ASSD}{average symmetric surface distance}
\newacronym{sgd}{SGD}{stochastic gradient descent}
\newacronym{miccai}{MICCAI}{International Conference on Medical Image Computing and Computer Assisted Intervention}
\newacronym[plural=OARs,firstplural=organs-at-risks (OARs)]{oar}{OAR}{organs-at-risk}
\newacronym[plural=GANs,firstplural=Generative Adversarial Networks (GANs)]{gan}{GAN}{Generative Adversarial Network}
\newacronym{ram}{RAM}{random access memory}

\newacronym{dct}{DCT}{Discrete Cosine Transform}
\newacronym{dwt}{DWT}{Discrete Wavelet Transform}
\newacronym{cwt}{CWT}{Continuous Wavelet Transform}
\newacronym{wt}{WT}{Wavelet Transform}
\newacronym{ft}{FT}{Fourier Transform}
\newacronym{hmt}{HMT}{Hidden Markov Tree}
\newacronym{lp}{LP}{Laplacian Pyramid}
\newacronym{dfb}{DFB}{Directional Filter Bank}  
\newacronym{pdfb}{PDFB}{Pyramid Directional Filter Bank}
\newacronym{ll}{LL}{Low Low}
\newacronym{lh}{LH}{Low High}
\newacronym{hl}{HL}{High Low}
\newacronym{hh}{HH}{High High}

\newacronym{asd}{ASD}{symmetric surface distance}
\newacronym{tpe}{TPE}{Tree-structured Parzen Estimator Approach}
\newacronym{rmse}{RMSE}{root mean square error}
\newacronym{ssl}{SSL}{semi-supervised learning}

\newacronym{pann}{PaNN}{Prior-aware Neural Network}

\newacronym{2D}{2D}{two-dimensional}
\newacronym{3D}{3D}{three-dimensional}
\newacronym{adain}{AdaIN}{adaptive instance normalization}
\newacronym{fid}{FID}{Fr\'{e}chet Inception Distance}

\newacronym{bsg}{B-SG}{Baseline StyleGAN2}
\newacronym{msg}{M-SG}{Mask StyleGAN2}
\newacronym{csg}{C-SG}{Conditional StyleGAN2}
\newacronym{roi}{ROI}{region of interest}
\newacronym{fir}{FIR}{finite impulse response}

\newacronym{1d}{1D}{one-dimensional}
\newacronym{2d}{2D}{two-dimensional}
\newacronym{3d}{3D}{three-dimensional}

\newcommand{\ie}{\textit{i.e.}\xspace}

\newcommand{\eg}{\textit{e.g.}\xspace}

\DeclareMathOperator{\EX}{\mathbb{E}}
\newcommand{\sg}{StyleGAN}
\newcommand{\con}{conditional~}
\newcommand{\tabref}[1]{Table~\ref{#1}}

\DeclareMathOperator{\gen}{G}
\DeclareMathOperator{\dis}{D}

\usepackage{amsmath,amsfonts,bm}

\def\figref#1{figure~\ref{#1}}
\def\Figref#1{Figure~\ref{#1}}

\def\Secref#1{Section~\ref{#1}}

\def\eqref#1{equation~\ref{#1}}

\def\1{\bm{1}}

\def\vm{{\bm{m}}}

\def\vx{{\bm{x}}}

\def\vz{{\bm{z}}}

\def\mM{{\bm{M}}}

\DeclareMathAlphabet{\mathsfit}{\encodingdefault}{\sfdefault}{m}{sl}
\SetMathAlphabet{\mathsfit}{bold}{\encodingdefault}{\sfdefault}{bx}{n}

\begin{document}

\maketitle

\begin{abstract}
Recent years have witnessed a growing academic and industrial interest in \gls{dl} for medical imaging. To perform well, \gls{dl} models require very large labeled datasets. However, most medical imaging datasets are small, with a limited number of annotated samples. The reason they are small is usually because delineating medical images is time-consuming and demanding for oncologists. There are various techniques that can be used to augment a dataset, for example, to apply affine transformations or elastic transformations to available images, or to add synthetic images generated by a \gls{gan}. In this work, we have developed a novel conditional variant of a current \gls{gan} method, the StyleGAN2, to generate multi-modal high-resolution medical images with the purpose to augment small medical imaging datasets with these synthetic images. We use the synthetic and real images from six datasets to train models for the downstream task of semantic segmentation. The quality of the generated medical images and the effect of this augmentation on the segmentation performance were evaluated afterward. Finally, the results indicate that the downstream segmentation models did not benefit from the generated images. Further work and analyses are required to establish how this augmentation affects the segmentation performance.
\end{abstract}

\glsresetall
\section{Introduction}

There has been a recent surge in the interest in \gls{dl} for medical imaging. This is partly due to the advancements in computing power (and storage), the availability of large datasets, and recent methodological and technical developments in the area. Computer vision, speech recognition, natural language processing, and bioinformatics are a few areas where \gls{dl} has been put to use, with results obtained that are on par with, and in some cases better than, those achieved by human experts~\citep{kuhl2022human}. \Gls{dl} has been used widely in the medical imaging field since \gls{dl} models can be used to perform diagnosis and other tasks by directly interpreting medical images.

Deep generative models are a particular type of \gls{dl} models that can learn a distribution over arbitrary data, \eg~over medical images. A \gls{gan} is a development in deep generative modeling~\citep{goodfellow2014generative} that has seen substantial interest in the last years. In this work, we explore the potential of using \glspl{gan} to augment small medical image datasets with synthetic images for the task of medical image segmentation.


A \gls{gan} model contains two (neural network) models, a generator and a discriminator~\citep{goodfellow2014generative}. The generator attempts to generate synthetic but realistic samples such that the discriminator will mistake these generated samples for real samples. The discriminator is, on the other hand, trained to determine whether a given sample is generated or actual real data. In other words, the generator is not tasked with minimizing some distance to a target image or so but is instead trained to fool the discriminator. This makes it possible for the model to learn without supervision. The generator and the discriminator are trained in an alternating min-max procedure where the generator tries to minimize the discriminator's ability to distinguish generated images from real, and the discriminator attempts to maximize its own ability to discriminate generated samples from real. The distribution of the generated samples should ideally be identical to the training data distribution after training, and it can be shown that the min-max training procedure can achieve this~\citep{goodfellow2014generative}.
In most cases, the generator is the primary focus, and the discriminator is often discarded once the generator has been trained.

Intriguing \gls{gan} applications include generating images in the style of another image~\citep[image-to-image translation,][]{han2018spine}, image inpainting~\citep[filling in missing parts of an image,][]{armanious2019adversarial}, and data augmentation~\citep[generating synthetic training data,][]{calimeri2017biomedical,madani2018chest,bowles2018gan,kim2021synthesis,motamed2021data,sandfort2019data}.

One of the main problems in medical imaging is having small datasets, with a limited number of annotated samples. This problem is emphasized in applications with \gls{dl} since \gls{dl} models need large labeled datasets. Experts, with extensive expertise in the data and the task at hand, often create annotations for medical imaging projects. It is often time-consuming to make these medical image annotations, particularly so for detailed annotations, such as segmentations of organs or lesions in many 2D slices of a 3D volume. Even though medical datasets are publicly accessible online, the majority of datasets are still limited in size and are only relevant for the specific medical concerns it was collected for.

The use of data augmentation~ is one strategy that researchers are pursuing to address this dilemma~\citep[\ie, the dilemma of having datasets with limited sizes,][]{shorten2019survey}. The most used data augmentation techniques involve applying various spatial transformations to the original dataset, such as translation, rotation, flip, and scale, or color transformations, such as randomly changing an image's hue, saturation, brightness, and contrast. These methods take advantage of changes that we know should not change the image's assigned class. This method is very effective for small datasets. Even models trained on some of the largest publicly available datasets, such as the ImageNet dataset~\citep{deng2009imagenet}, can benefit from this method~\citep{shorten2019survey}. In computer vision tasks, it is customary to use conventional data augmentation to assist the training process of neural network models. However, small transformations of images may not add much extra information (\eg, the translation of an image by a few pixels). A novel and advanced data augmentation technique is to do data augmentation with high-quality and high-resolution generated synthetic images~\citep{karras2018progressive,karras2018stylegan,Karras2019stylegan2}. Synthetic images, or other data, created using a generative model may provide data with greater diversity that enriches the training data and improve the final model.

The main contributions of this work can be summarized as follows:
\begin{enumerate}
    \item We propose a novel conditional \gls{gan} based on the \sg2 model and refer to it as a ``Conditional-\sg2 model''.
    \item We have systematically investigated the \sg2~\citep{Karras2019stylegan2} and the proposed Conditional-\sg2. 
\end{enumerate}

\section{Related Work}

It is challenging to generate images with a high resolution; because the higher the resolution of an image, the easier it is for the discriminator to distinguish between ``fake'' and ``real'' images. Smaller mini-batches, that decrease the stability of the model training, and memory limits on commodity \glspl{gpu} are also two problems of generating high-resolution images~\citep{odena2017conditional}. 
\citet{karras2018progressive} addressed this issue by proposing the progressive growing of \gls{gan} (Progressive-\gls{gan}). Their key contribution was to progressively increase the resolution of both the generator output and the discriminator input, beginning with low-resolution images and successively adding new and larger layers that produce higher-resolution images. This significantly improved the generated image quality, accelerated the training, and improved the stability of training when generating high-resolution images. \citet{karras2018stylegan} later proposed the \sg{} as a combination of Progressive-\gls{gan}~\citep{karras2018progressive} and what's called neural style transfer~\citep[\ie, to copy the style of a source image to a new target image,][]{gatys2016neural}. A standard generator takes a latent code (a random noise vector) as an input, but \citet{karras2018stylegan} also proposed to map this latent input to an intermediate latent space that has fewer problems of entanglement between variables in the input latent space---\ie, dimensions in the intermediate space are more likely to encode only a single feature. The intermediate latent vectors are then fed into a generator at multiple positions.


\begin{figure*}[!th]
    \centering
    \includegraphics[width=.9\textwidth]{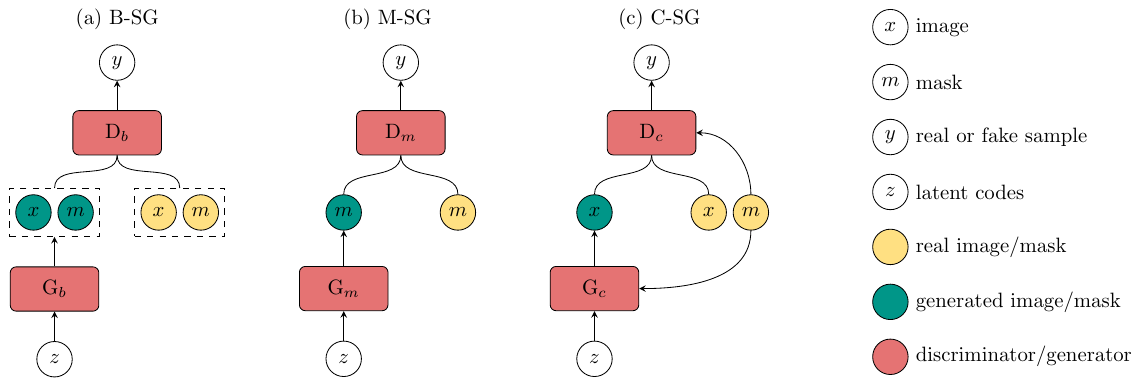}
    \caption{Comparison of \glspl{gan} used in this work.}
    \label{fig:proposed_gan}
\end{figure*}

\citet{Karras2019stylegan2} further proposed the \sg2, an improvement that restructures the normalization of the generator and also regularizes the generator. The restructuring improved the generated images' quality, and their \textit{path length regularizer} made the generator simpler to invert (to find the latent code that corresponds to an image).
In follow-up work, \citet{karras2020training} proposed the \sg2-ADA to address the problem of overfitting the discriminator, which causes training to diverge due to insufficient data. In that work, they presented an adaptive discriminator augmentation approach that controls the stability of training on datasets with limited data. The \sg2-ADA did not introduce any other modifications to the loss functions or the network designs. In a recent work, \citet{karras2021alias} introduced the \sg3, that eliminates \textit{texture sticking}, induced by point-wise nonlinearities. Texture sticking happens when a generator memorizes a certain area's textural features. To address the issue of texture sticking, \citet{karras2021alias} also proposed interpreting all network signals as continuous and applying low-pass filters on these signals.


The \con \gls{gan} was first introduced by \citet{mirza2014conditional} as a conditional version of the original \gls{gan}~\citep{goodfellow2014generative}. The idea is to generate images given some piece of information, such as a desired label. Practically, this was achieved by passing the desired label as an input to the generator and the discriminator to inform about what was to be generated. \citet{isola2017image} extended the idea of a \con \gls{gan} by conditioning on a source image to generate a target image and presented an image-to-image translation model called pix2pix (for pixel-to-pixel translation). They demonstrated that the pix2pix model is effective for various image translation applications, and one of these was image synthesis from binary label maps. \citet{bailo2019red} applied image-to-image translation on blood smear data to generate new samples and substantially expanded their small datasets. In particular, given the mask of a microscopy image, they generated high-quality blood cell samples, which were then used in conjunction with real data during network training for segmentation and object detection tasks.

\Glspl{gan} have been utilized for data augmentation to improve the training of \glspl{cnn} by generating additional samples from the training data distribution. \citet{calimeri2017biomedical} proposed to use a \gls{gan} to generate \gls{mri} slices of the human brain. In another work, \citet{madani2018chest} investigated using \glspl{gan} to augment chest X-ray images that were then used when training a \gls{cnn} for cardiovascular abnormality classification. \citet{sandfort2019data} used CycleGAN~\citep{zhu2017unpaired} to generate contrast \gls{ct} from non-contrast \gls{ct} images. In all these works above, they showed that using \gls{gan} to generate additional data for augmentation can provide significant improvements to the downstream task (\eg, for segmentation or classification). \citet{skandarani2021gans}, however, argued that no \gls{gan} is capable of reproducing the full richness of medical datasets. In specific, despite being trained on a significantly larger number of samples (real and generated), none of the segmentation networks' performance match or surpass those trained on only the real data.


\section{Proposed Methods}

\glspl{gan} are generative models that map a random noise vector, $\vz$, to an output sample (\eg, an image). Depending on the problem, the output image can be a gray image~\citep{goodfellow2014generative}, a color image~\citep{Karras2019stylegan2}, or a multi-modal medical images~\citep{zhu2017unpaired}. The primary aim of this paper is to detail a proposition to use \glspl{gan} to generate medical images and their corresponding label masks. We trained \glspl{gan} on an existing dataset and used the generated synthetic samples, pairs that included both medical images (single-modal or multi-modal) and their masks, to enlarge the dataset for the downstream task of semantic segmentation. We trained a U-Net~\citep{ronneberger2015unet} on the real samples and generated samples for the downstream task. We hypothesized that the segmentation performance when training on real and augmented samples (\ie, generated images and masks) would be higher than when training on only real samples.

\subsection{Objective}

In this work, we propose two \glspl{gan} based on the \sg2~\citep{Karras2019stylegan2}. The first \sg2 was trained to generate label masks and is referred to as \gls{msg}. The second \gls{gan} was a conditional variant of \sg2 and is referred to as \gls{csg}. The condition for \gls{csg} was the label mask. At the same time, the output of \gls{csg}'s generator was that mask's corresponding medical images. Third, we also use a standard, or baseline \sg2, referred to as \gls{bsg}, to generate a pair of medical images and label masks. We considered the \gls{bsg} as the baseline \gls{gan} model. \Figref{fig:proposed_gan} illustrates the three \gls{gan} models considered in this work. We detail the three \glspl{gan} below.

\begin{figure*}[!th]
\centering
\includegraphics[width=.9\textwidth]{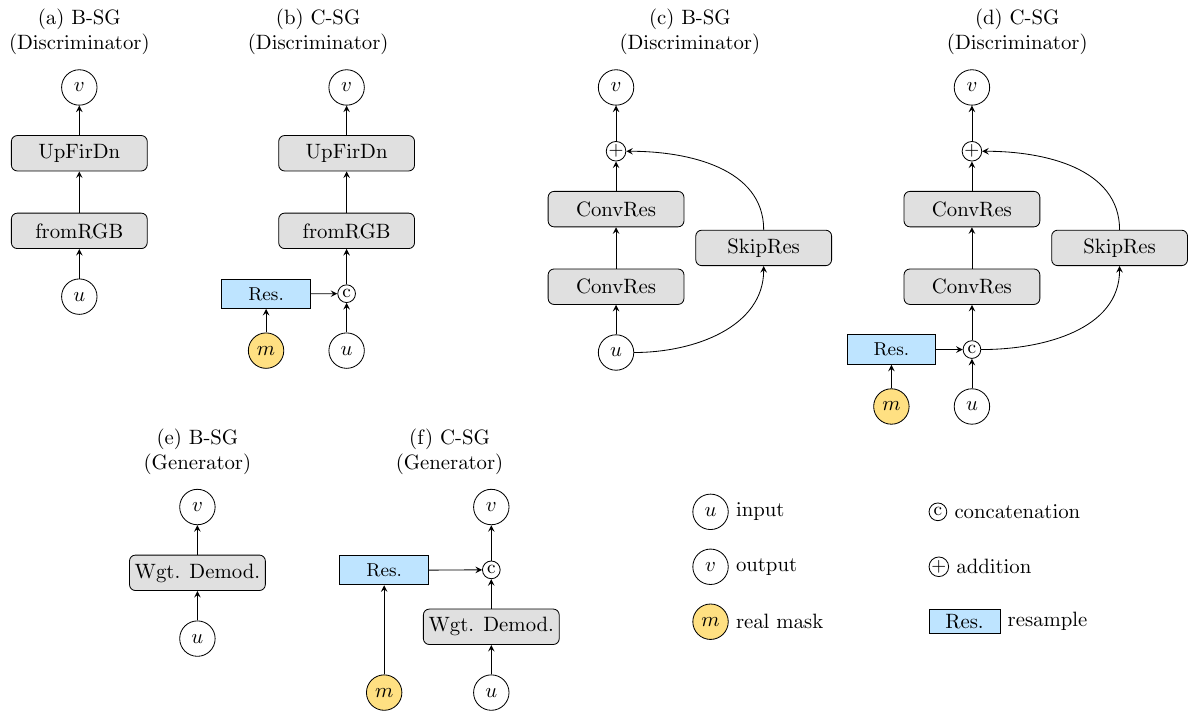}
\caption{Comparison of architectures of \gls{bsg} (or \gls{msg}) and \gls{csg}. The ``Wgt. Demod.'' denote the weight demodulation module proposed by~\citet{Karras2019stylegan2}.}
\label{fig:proposed_csg}
\end{figure*}

Let the training dataset be
$$
    \mathbf{F} 
        = 
            \big\{ 
            (\vx_{i}, \vm_{i}) : \forall i \in \{1, \dots, N\} 
            \big\},
$$
where $N$ is the number of training pairs. The $(\vx_{i}, \vm_{i})$ denotes the $i$-th training pair with an image, $\vx_i$, and its corresponding label mask, $\vm_i$.

We construct the label mask training set from $\mathbf{F}$ as
$$
    \mathbf{M} 
        = 
            \big\{ 
            \vm_{i} :
                (\vx_{i}, \vm_{i}) \in \mathbf{F}
            \big\}.
$$

For the \gls{bsg}, the objective function of a two-player min-max game~\citep{goodfellow2014generative} can be defined as
\begin{align}
    \min_{\gen_b} \max_{\dis_b} V(\gen_b, \dis_b) 
        & = 
            \EX_{(\vx, \vm) \sim p_{\mathbf{F}}(\vx, \vm)}
            [\log \dis_b(\vx, \vm)] 
            \nonumber \\ 
        & \quad + 
            \EX_{\vz \sim p_{\vz}(\vz)}[\log (1 - \dis_b(\gen_b(\vz)))].
            \label{eq:minimaxgame-bsg}
\end{align}
with $\gen_b$ and $\dis_b$ the generator and discriminator of the \gls{bsg}. The $p_{\mathbf{F}}(\vx, \vm)$ denotes the underlying data distribution of the training data in $\mathbf{F}$, while $p_{\vz}(\vz)$ is a prior on the latent noise vectors.


The objective function of the proposed \gls{msg} is also a two-player min-max game and can be written as
\begin{align}
    \min_{\gen_m} \max_{\dis_m} V(\gen_m, \dis_m) 
        & = 
            \EX_{\vm \sim p_{\mathbf{M}}(\vm)}
            [\log \dis_m(\vm)] 
            \nonumber \\ 
        & \quad +         
            \EX_{\vz \sim p_{\vz}(\vz)}[\log (1 - \dis_m(\gen_m(\vz)))],
            \label{eq:minimaxgame-msg}
\end{align}
with $\gen_m$ and $\dis_m$ the generator and discriminator of the \gls{msg}. The $p_{\mathbf{M}}(\vm)$ denotes the underlying data distribution of training masks in $\mathbf{M}$, and $p_{\vz}(\vz)$ is again a prior on the latent noise vectors.

Furthermore, we define the objective function of the proposed \gls{csg} as
\begin{align}
    \min_{\gen_c} \max_{\dis_c} V(\gen_c, \dis_c) 
        & = 
            \EX_{(\vx, \vm) \sim p_{\mathbf{F}}(\vx, \vm)}
            [\log \dis_c(\vx|\vm)] 
            \nonumber \\ 
        & \quad + 
            \EX_{\vz \sim p_{\vz}(\vz)}[\log (1 - \dis_c(\gen_c(\vz|\vm)))],
            \label{eq:minimaxgame-csg}
\end{align}
with $\gen_c$ and $\dis_c$ the generator and discriminator of the \gls{csg}.

\subsection{\acrlong{csg} (\gls{csg})}

Basically, \gls{bsg} and \gls{msg} use the same architecture as the \sg2. The only difference between \gls{bsg} and \gls{msg} is the output of the generator, \ie the output of \gls{bsg} is pairs of both the label masks and the medical images. In contrast, the output of \gls{msg} is only the label masks.

\begin{figure*}[!th]
    \centering
    \includegraphics[width=\textwidth]{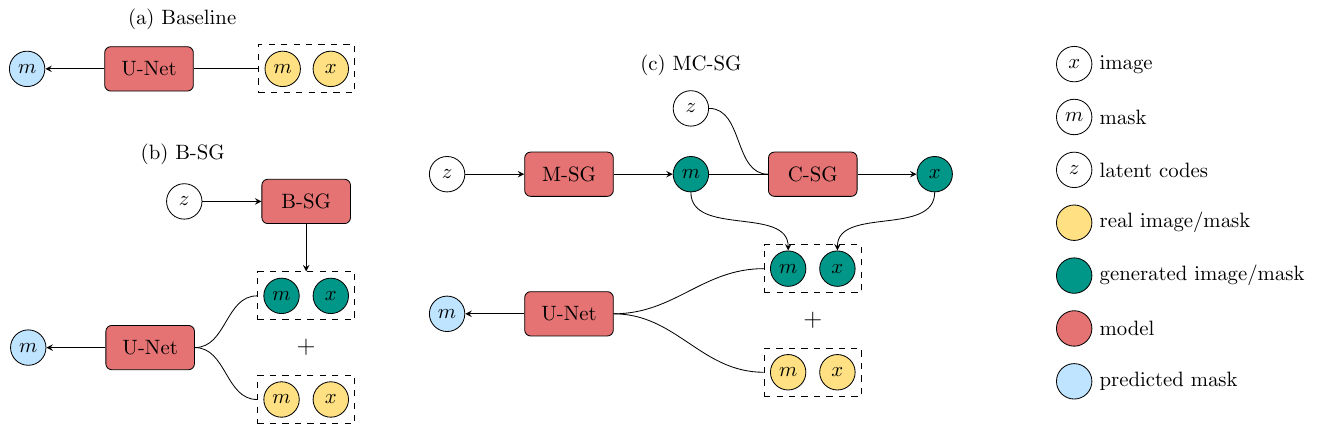}
    \caption{Proposed pipeline.}
    \label{fig:proposed_pipeline}
\end{figure*}

In order to use a condition as a mask in \sg2, we introduce the \gls{csg}. \Figref{fig:proposed_csg} shows a comparison between the design of the \gls{bsg} (also \gls{msg}) and of the proposed \gls{csg}. From \Figref{fig:proposed_csg}, we can see that the \gls{csg} differs the \gls{bsg} on three primary blocks: two in the discriminator (\Figref{fig:proposed_csg}(a)--\Figref{fig:proposed_csg}(d)) and one in the generator (\Figref{fig:proposed_csg}(e)--\Figref{fig:proposed_csg}(f)).

To incorporate the condition of the mask on the generated output in the generator, we feed a mask into the generator (see \Figref{fig:proposed_csg}(f)). In the \gls{csg}, we resampled the label mask to the shape of the output of the weight demodulation module proposed by \citet{Karras2019stylegan2}. The resampled mask is then concatenated with the output of the weight demodulation module.

In the discriminator of \gls{csg}, we provided the mask condition at two locations (see \Figref{fig:proposed_csg}(b) and \Figref{fig:proposed_csg}(d)). First, we concatenated the resampled mask with the input before it was fed into a layer called ``fromRGB''. The ``fromRGB'' is a layer that converts an RGB image to a feature map. This layer, together with ``toRGB'', that converts a feature map to an RGB image, was first introduced for the Progressive-\gls{gan}~\citep{karras2018progressive}. The output of the concatenation operation was then fed into a block named ``UpFirDn''. The ``UpFirDn'' block includes a padding operation, an upsampling operation, a filtering operation with a given \gls{fir} filter, and followed by a downsampling operation. Second, we concatenated the resampled mask with the input before it was fed into a ResNet-like block~~\citep[see \Figref{fig:proposed_csg}(d),][]{He2016Deepa}. The ResNet-like block consisted of six components: two convolutional layers followed by two resampling layers and a skip connection followed by a resampling layer.

\subsection{Proposed Pipeline} \label{sec:proposed_pipeline}

In this work, the downstream task is the semantic segmentation of medical images. The segmentation network used in this work was the U-Net~\citep{ronneberger2015unet}. We split each dataset into three parts: training (60\%), validation (20\%), and test (20\%). \Figref{fig:proposed_pipeline} shows a comparison of three segmentation pipelines as follows. First, the baseline (BL) that utilized the U-Net to train only on the real samples (see \Figref{fig:proposed_pipeline}(a)). Second, the B-SG generated pairs of images and label masks. After that, the generated and the real samples were used to train the U-Net (see \Figref{fig:proposed_pipeline}(b)). Third, the MC-SG is the proposed pipeline (see \Figref{fig:proposed_pipeline}(c)). We first used \gls{msg} to generate masks. We then used the generated masks by \gls{msg} as conditions to make \gls{csg} generate images. The generated samples from MC-SG were then employed with the real samples to train another U-Net for the downstream task.


\subsection{Statistical Tests} \label{sec:stat-tests}

We employed the Friedman test of equivalence on the evaluated metrics. It was used with the predictions made on the test set to compare the performances of the evaluated methods. Following~\citet{Demsar2006}, we also detailed the significant differences between each pair of methods (\ie better, worse, or undetermined) by a Nemenyi post-hoc test of pair-wise differences.

\section{Experiments} \label{sec:exp}


\subsection{Datasets} \label{sec:dataset}

The experiment was conducted on six datasets, spanning a range of anatomies, medical imaging modalities, and dataset sizes, in order to evaluate the capabilities of the proposed method. Five of the datasets are accessible to the public as segmentation challenges. In addition to these public datasets, we made use of an in-house dataset collected at the University Hospital of Ume\aa{}, Ume\aa{}, Sweden.

\begin{table*}[!th] 
	\def\widthdetail{2.2cm}
	\def\width{3.2cm}
	\caption{The datasets, augmentation, and experimental setup in this study. A depth of $D$ means that the volume shape had a varied depth.}
	\centering
         \begin{adjustbox}{max width=\textwidth}
	\begin{tabular}{@{}L{\widthdetail} l R{\width} R{\width} R{\width} R{\width} R{\width} R{\width} R{\width}@{}}
		\toprule
		material/dataset                &  & \acrshort{brats}  	& \acrshort{kits}	& \acrshort{ibsr}   & \acrshort{pros}	& \acrshort{heart}	& \acrshort{spleen}	\\ 
		\midrule
		type                            &  & \acrshort{mri}     & \acrshort{ct}     & \acrshort{mri}    & \acrshort{ct}     & \acrshort{mri}    & \acrshort{ct}     \\ 
		\#modalities                    &  & 4                  & 1                 & 1                 & 1                 & 1                 & 1                 \\
		\#classes                       &  & 3                  & 2                 & 3                 & 3                 & 1                 & 1                 \\
		\midrule
		\#patients                      &  & 369                & 210               & 18                & 500               &  20               &  41               \\
		\#patients cases         		&  & $\{20,50,100,200,369\}$  
																& $\{20,50,100,210\}$ 
																				  & $\{18\}$          & $\{20,50,100,200,500\}$
																														  &  $\{20\}$
																																				&  $\{20,41\}$      \\		
		\midrule
		original shape                  &  &  240-240-155       &  512-512-$D$      &  256-128-256      &  512-512-$D$      &  320-320-$D$      &  512-512-$D$      \\
		resized shape                   &  &  256-256-155       &  256-256-$D$      &  256-256-256      &  256-256-$D$      &  256-256-$D$      &  256-256-$D$      \\
		\midrule
		\multicolumn{7}{@{}l}{augmentation}  \\
		\midrule
		flip left-right         		&  & \cmark         	& \xmark        	& \cmark        	& \cmark        	& \xmark        	& \xmark        	\\
		rotation                		&  & \cmark         	& \cmark        	& \cmark        	& \cmark        	& \cmark        	& \cmark        	\\
		shift                   		&  & \cmark         	& \cmark        	& \cmark        	& \cmark        	& \cmark        	& \cmark        	\\
		zoom                    		&  & \cmark         	& \cmark        	& \cmark        	& \cmark        	& \cmark        	& \cmark        	\\
		\midrule
		\multicolumn{7}{@{}l}{training}  \\
		\midrule
		\#epochs            			&  & 70                 & 70                & 150               & 70                & 150             	& 150               \\
		optimizer           			&  & Adam               & Adam              & Adam              & Adam              & Adam              & Adam              \\
		learning rate       			&  & $1\cdot10^{-4}$    & $1\cdot10^{-4}$   & $1\cdot10^{-4}$   & $1\cdot10^{-4}$   & $1\cdot10^{-4}$ 	& $1\cdot10^{-4}$   \\
		batch-size            			&  & 64                 & 64                & 64                & 64                & 64              	& 64               \\
		\bottomrule
	\end{tabular}
        \end{adjustbox}
	\label{tab:datasetinfo}
\end{table*}

\textit{The \gls{pros}} is an in-house dataset that contains \gls{ct} scans of the pelvis region from $1\,244$ patients who had radiotherapy treatments for prostate cancer at the University Hospital of Ume\aa{}, Ume\aa{}, Sweden. The bladder and the rectum are two of the organs that are considered to be at risk. In addition to the bladder and rectum, the delineations also include the prostate, which is most of the time referred to as the clinical target volume. The separate structure masks for these three aforementioned structures were combined into a single mask image, with the pixel value of $1$ assigned to the prostate, $2$ assigned to the bladder, and $3$ assigned to the rectum. Patients who did not have all three structures were removed, leaving $1\,148$ patients. We then randomly selected $500$ patients for this study.

In connection to the \gls{miccai} main event that took place in 2020, the \textit{\gls{brats}}~\citep{menze2014multimodal,bakas2017advancing} was organized as a satellite event. The \gls{brats} dataset contains $369$ volumes of \gls{3D} multiple pre-operative \gls{mri} from 19 different hospitals that have either \gls{hgg} or \gls{lgg}. The following \gls{mri} sequences were performed on each patient: \gls{t1}, \gls{t1c}, \gls{t2}, and \gls{flair}. These scans were performed utilizing various techniques and scanners, and the magnetic field strength was set to $3$T. Several annotators were present to do manual segmentations. A contrast-enhancing tumor was differentiated from a necrotic and non-enhancing tumor core, as well as peritumoral edema. The skulls have been cropped out of the images. After that, the voxels were interpolated to a single size and co-registered to a single anatomical reference.

The \textit{\gls{ibsr}} dataset~\citep{cocosco1997brainweb}  was released by the National Institute of Neurological Disorders and Stroke (NINDS). The dataset contains $18$ \acrlong{3D} T1-weighted \gls{mri} scans of 1.5mm slice thickness. Radiologists annotated the contours of the whole brain, including cerebrospinal fluid (label 1), gray matter (label 2), and white matter (label 3). 

The \textit{\gls{kits}} was a grand challenge in conjunction with the \gls{miccai} in 2019, which was held in Shenzhen, China~\citep{heller2019kits19}. The challenge aimed to autonomously segment contrast-enhanced \gls{ct} images of the abdomen into three classes: kidney, tumor, and background. The patients included in this dataset were chosen randomly from those who had undergone radical nephrectomy at the University of Minnesota Medical Center between 2010 and 2018.

The \textit{\gls{spleen}} dataset is one out of ten labeled datasets provided by the Medical Segmentation Decathlon that covers a wide range of organs~\citep{simpson2019large}. The \gls{spleen} dataset's \gls{roi} target is the spleen. The dataset initially belonged to a study on splenic volume change following chemotherapy in patients with liver metastases and consists of $41$ \gls{ct} scans with annotated masks. 

The \textit{\gls{heart}} is a small dataset and is also part of Medical Segmentation Decathlon~\citep{simpson2019large}. The \gls{heart} dataset contains $20$ mono-modal \gls{mri} scans with annotated masks. The \gls{roi} target of this dataset is the left atrium.

\subsection{Experiments} \label{sec:experiments}

We carried out the experiments on six datasets (see \Secref{sec:dataset}). To evaluate the capability of evaluated methods on diverse datasets sizes, we randomly selected several number of samples from each dataset to form new sub-dasesets with specific sizes. Row ``\#patients cases'' in \tabref{tab:datasetinfo} shows the numbers of patients that were randomly selected. For example, the ${20, 41}$ of the \gls{spleen} means that we created two distinguish sub-datasets from the \gls{spleen} dataset: one dataset had 20 samples, and the other had 41 samples. 

In order to evaluate the effect of the fraction of generated training samples for the segmentation task, we trained multiple U-Nets for the different number of generated samples (by the \gls{bsg} and the MC-SG, see \Secref{sec:proposed_pipeline}).
The numbers of generated samples were 20, 50, 100, 200, 400, and 500 patients. 
Consequently, each pipeline that used generated samples from \gls{gan} had 6 variants, and each variant corresponded to a number of generated patients.

We also applied simple data augmentation techniques (spatial transformations) to each U-Net model. Hence, we ended up with 26 methods in total, \ie $(\text{BL} + \text{6 variants of B-SG} + \text{6 variants of MC-SG}) \times (\text{2 data augmentation})$, see \tabref{tab:result_nemenyi}.

\subsection{Evaluation} \label{sec:eval}

To measure the difference between two Gaussian distributions,~\citet{frechet1957distance} presented the Fr\'{e}chet distance. Inspired by that work,~\citet{heusel2017gans} proposed \gls{fid} to evaluate the quality of generated samples of \glspl{gan}. In this work, we used \gls{fid} to evaluate the \glspl{gan}. The \gls{fid} is computed as
\begin{align}
    d^2((\boldsymbol{\mu},\boldsymbol{\Sigma}),(\boldsymbol{\mu}_g,\boldsymbol{\Sigma}_g))
        & =
            \lVert \boldsymbol{\mu}-\boldsymbol{\mu}_g \rVert_2^2
            \nonumber \\
        & \quad +  
            \mathrm{tr} 
            \bigl(
            \boldsymbol{\Sigma}+\boldsymbol{\Sigma}_g-2\bigl(\boldsymbol{\Sigma}\boldsymbol{\Sigma}_g\bigr)^{1/2}
            \bigr),
\end{align} 
where $\mathrm{tr}$ denotes trace of a matrix. The $\lVert \cdot \rVert_2$ is the standard $\ell_2$ norm. The $d(\cdot,\cdot)$ denotes the Fr\'{e}chet distance between two Gaussian distributions. The $(\boldsymbol{\mu},\boldsymbol{\Sigma})$ and $(\boldsymbol{\mu}_g,\boldsymbol{\Sigma}_g)$ are the mean and covariance matrices of Gaussian distributions fitted to the feature representations of real and generated images, respectively. The feature representations ($2\,048$-length vectors)
is computed by feeding the images to an Inception-v3 network~\citep{szegedy2015inception} that was pre-trained on the ImageNet dataset~\citep{Krizhevsky2012Imagenet}. 

We used the soft \gls{dsc} loss in the downstream tasks (\ie segmentation). The soft \gls{dsc} loss has been used in many previous works, \eg by~\citet{vnet-dice,vu2020evaluation,vu2020tunet,vu2021multi,vu2021data}. The soft \gls{dsc} loss is computed as
\begin{equation} \label{eqn:dscloss}
    \mathcal{L}_{\mathrm{DSC}}(\mM, \hat{\mM}) 
        = 
            \frac{-2 \sum_{i=1}^N m_i \hat{m}_i + \epsilon}
            {\sum_{i=1}^N m_i + \sum_{i=1}^N \hat{m}_i + \epsilon},
\end{equation}
where $\hat{m}_i$ is the $i$-th element of the predicted softmax output of the network, and the $m_i$ is the $i$-th element of a one-hot encoding of the ground truth labels. To avoid division by zero, we added $\epsilon = 1 \cdot 10^{-5}$ to the denominator. We also added $\epsilon$ to the numerator to give true zero predicted masks a score of $1$

We evaluated the segmentation tasks using the \gls{dsc}, which is calculated as
\begin{equation}\label{eq:dice}
    \mathrm{DSC} (\mM, \hat{\mM}) 
        = 
            -\mathcal{L}_{\mathrm{DSC}}(\mM, \hat{\mM}).
\end{equation}
in which $\epsilon$ is set to zero.

\subsection{Implementation Details and Training} \label{sec:details}

The proposed method was implemented in PyTorch 1.10\footnote{\url{https://pytorch.org/}}. The experiments were run on NVIDIA Tesla K80 and V100 \glspl{gpu} housed at the \gls{hpc2n}\footnote{\url{https://www.hpc2n.umu.se/}} at Ume{\aa} University, Sweden.


We used the Adam optimizer~\citep{kingma2014adam} in all experiments, with a fixed learning rate of $1\cdot10^{-4}$ (see \tabref{tab:datasetinfo}). The number of epochs for each experiment was set at $70$ or $150$, \ie depending on the datasets. We used the \gls{2D} U-Net~\citep{ronneberger2015unet} for the segmentation tasks. Batch normalization was used after all convolutional layers: in the transition blocks and in the main network. The transition blocks (downsampling or upsampling) consist of a batch normalization layer, a \gls{relu} layer, and a convolution layer.



Except for the \gls{spleen}, all datasets were standardized to zero-mean and unit variance. Before normalization, N4ITK bias field correction~\citep{tustison2010n4itk} was applied to the \gls{brats} dataset. Samples from the \gls{spleen} dataset were rescaled to $[-1, 1]$.

To artificially enlarge the \gls{spleen} and \gls{brats} dataset size and increase the variability in the data, we applied different on-the-fly data augmentations. Those were: horizontal flipping, rotation within a range of $-1$ to $1$ degrees, rescaling with a factor of $0.9$ to $1.1$, and shifting the images by $-5$ to $5$ percent.

\section{Results and Discussion} \label{sec:results}







\begin{figure}[!th]
    \def\factor{1}
    \centering
	\includegraphics[width=\factor\columnwidth]{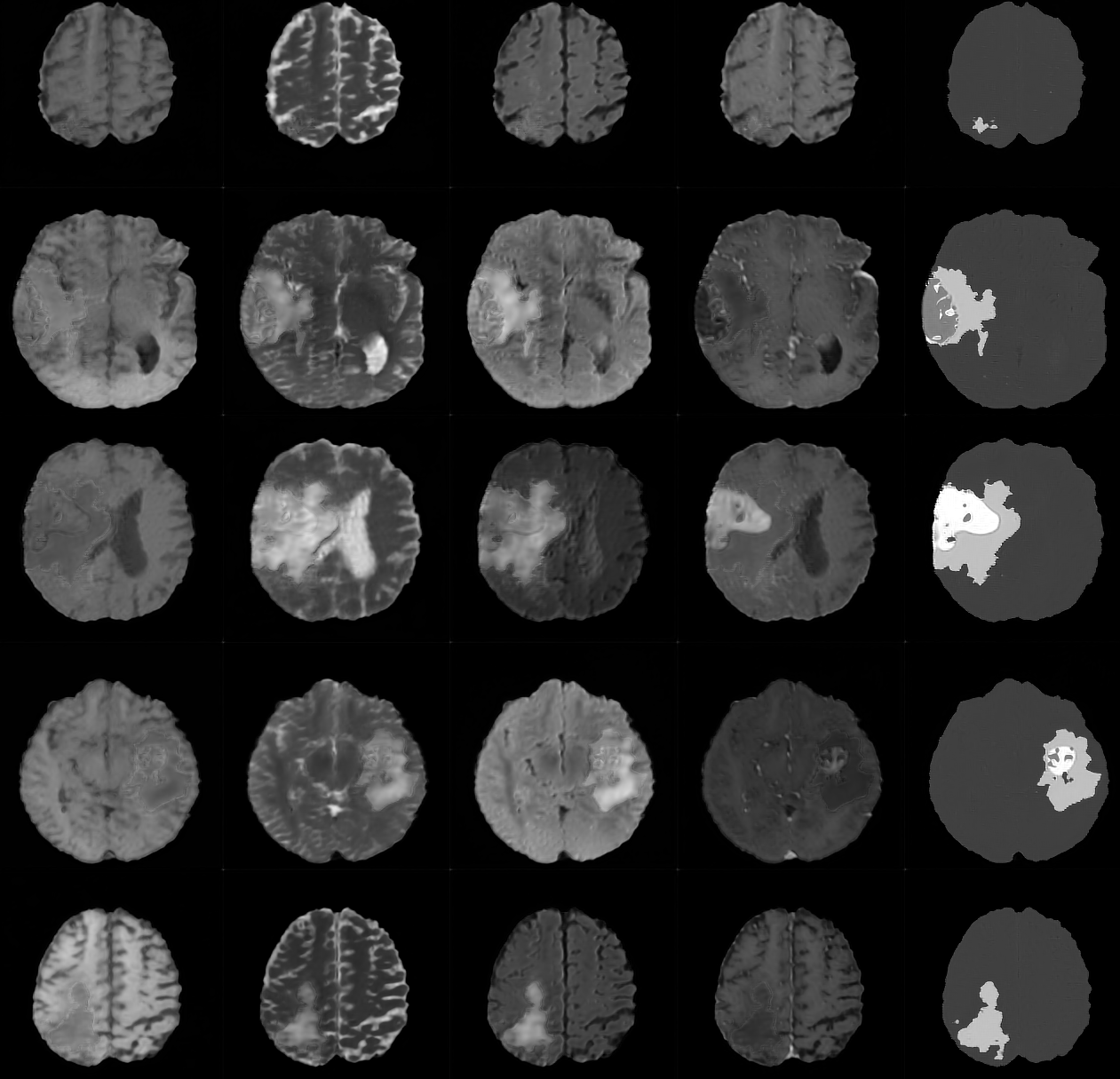}
    \caption{An uncurated set of generated samples from the \gls{brats} dataset using the generator of B-SG. From left to right: \gls{t1}, \gls{t2}, \gls{flair}, \gls{t1c} and the corresponding label mask.}
	\label{fig:gan_samples}
\end{figure}

\Figref{fig:gan_samples} shows an uncurated set of generated samples from the \gls{brats} dataset using the generator of B-SG. From left to right, the figure illustrates the \gls{t1}, \gls{t2}, \gls{flair}, \gls{t1c} and corresponding label mask. From \Figref{fig:gan_samples}, we can see that the average quality of the generated sample is not high. In specific, though the generated images have good structures, including lateral ventricles, third ventricle, thalamus, basal ganglia, brain lobes, tumor regions, and so on; however, when zooming in, we can see that the images look quite blotchy and blurry with low-quality.

\begin{figure*}[!th]
    \def\factor{0.19}
    \centering
    \begin{subfigure}[b]{\factor\textwidth}
		\centering
		\includegraphics[width=\textwidth]{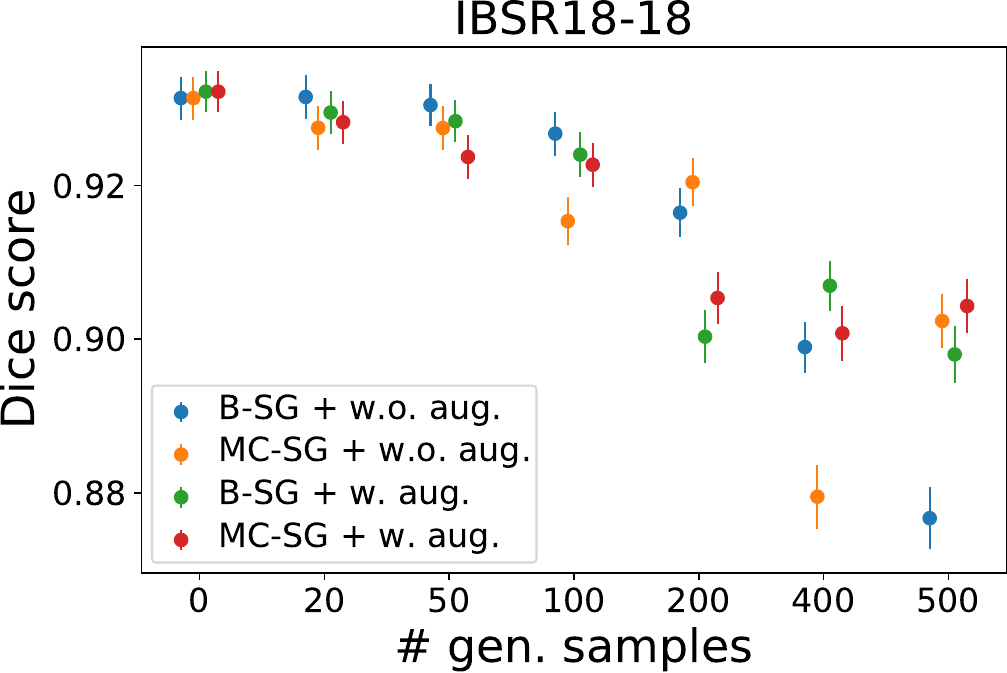}
    \end{subfigure}
    \begin{subfigure}[b]{\factor\textwidth}
		\centering
		\includegraphics[width=\textwidth]{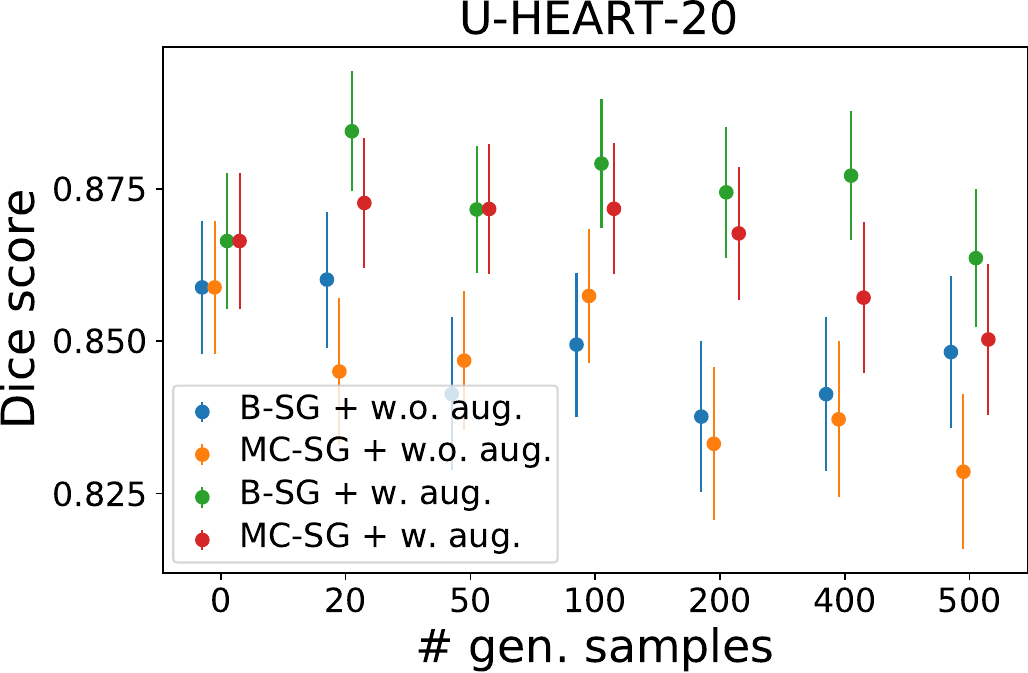}
    \end{subfigure}
    \begin{subfigure}[b]{\factor\textwidth}
		\centering
		\includegraphics[width=\textwidth]{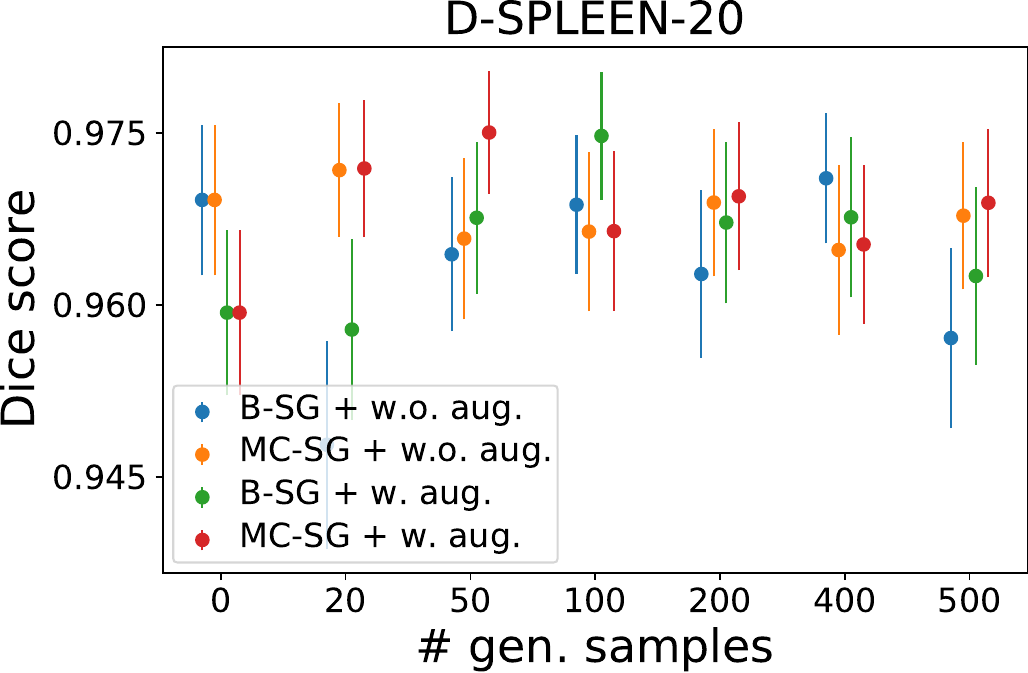}
    \end{subfigure}
    \begin{subfigure}[b]{\factor\textwidth}
		\centering
		\includegraphics[width=\textwidth]{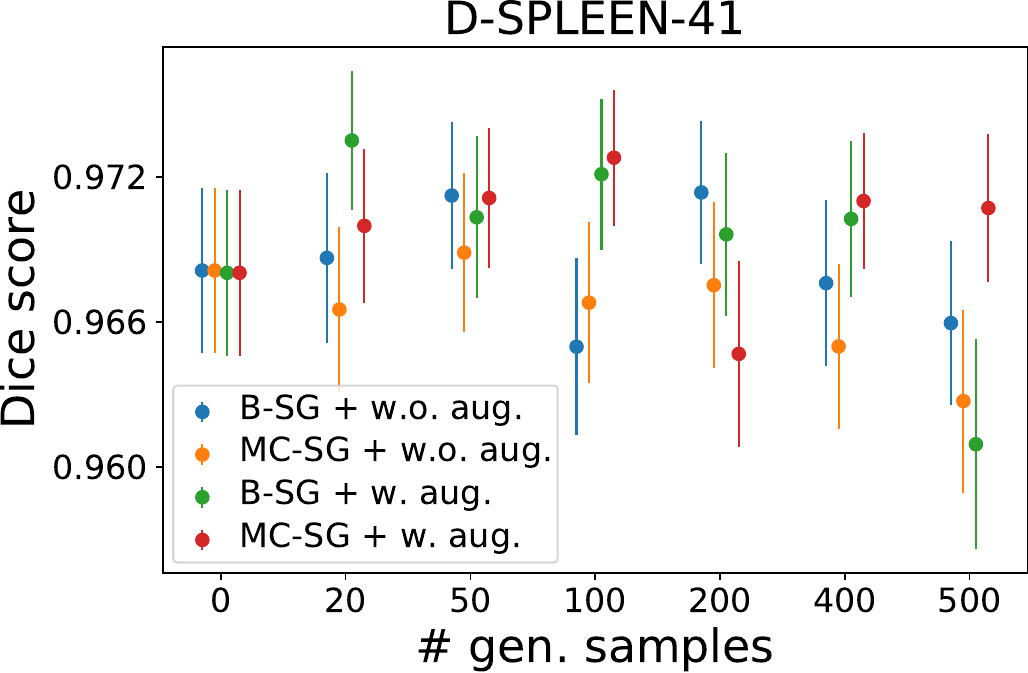}
    \end{subfigure} 
    \\
    \vspace{2mm}
    \begin{subfigure}[b]{\factor\textwidth}
		\centering
		\includegraphics[width=\textwidth]{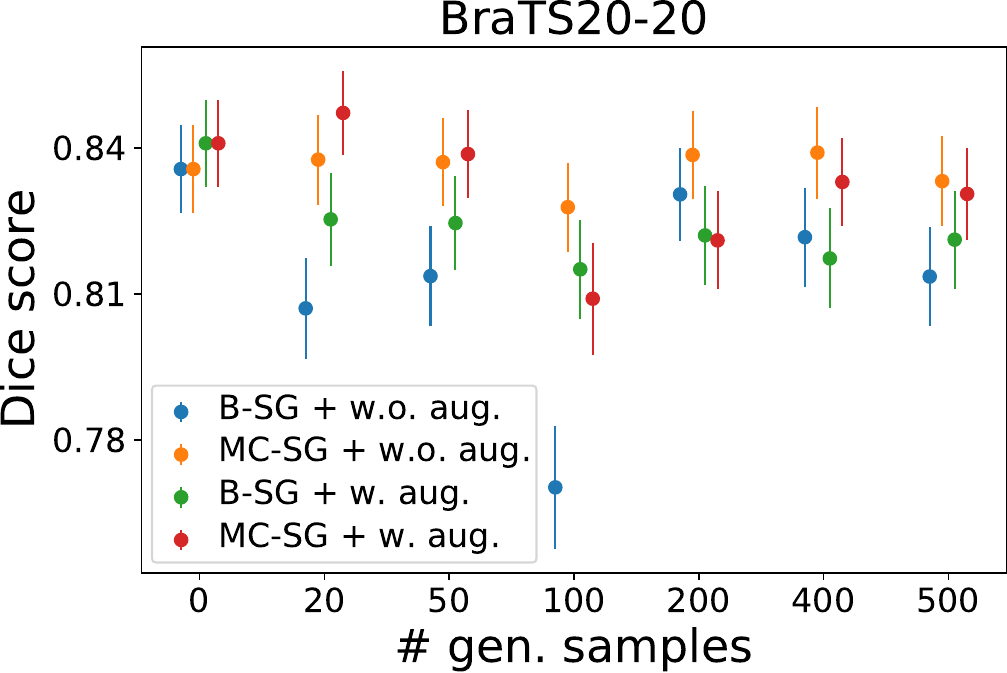}
    \end{subfigure}
    \begin{subfigure}[b]{\factor\textwidth}
		\centering
		\includegraphics[width=\textwidth]{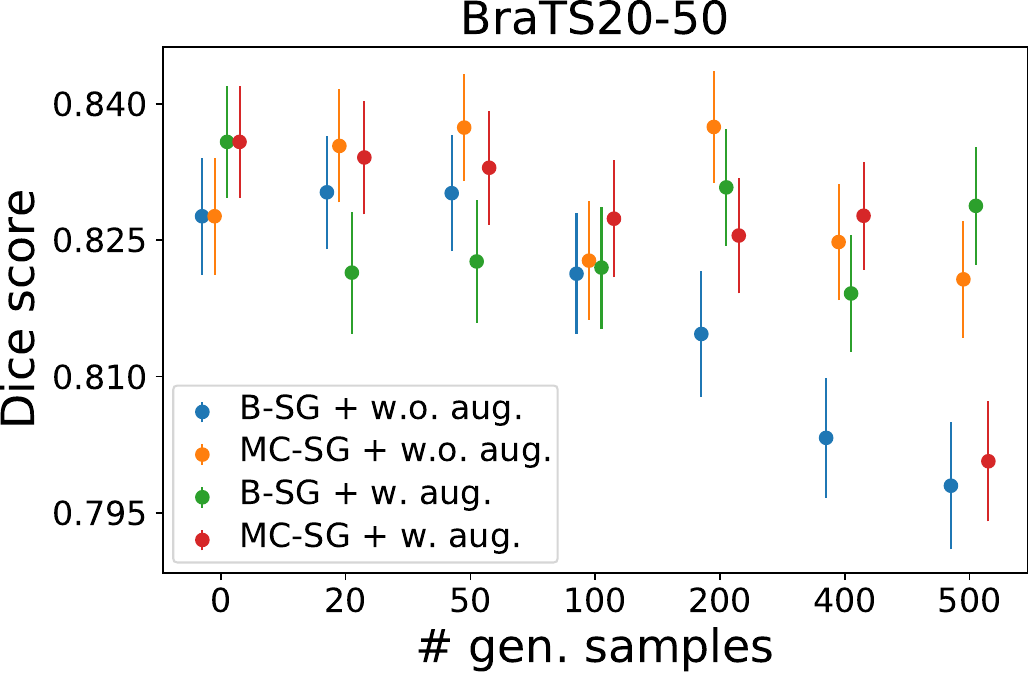}
    \end{subfigure}
    \begin{subfigure}[b]{\factor\textwidth}
		\centering
		\includegraphics[width=\textwidth]{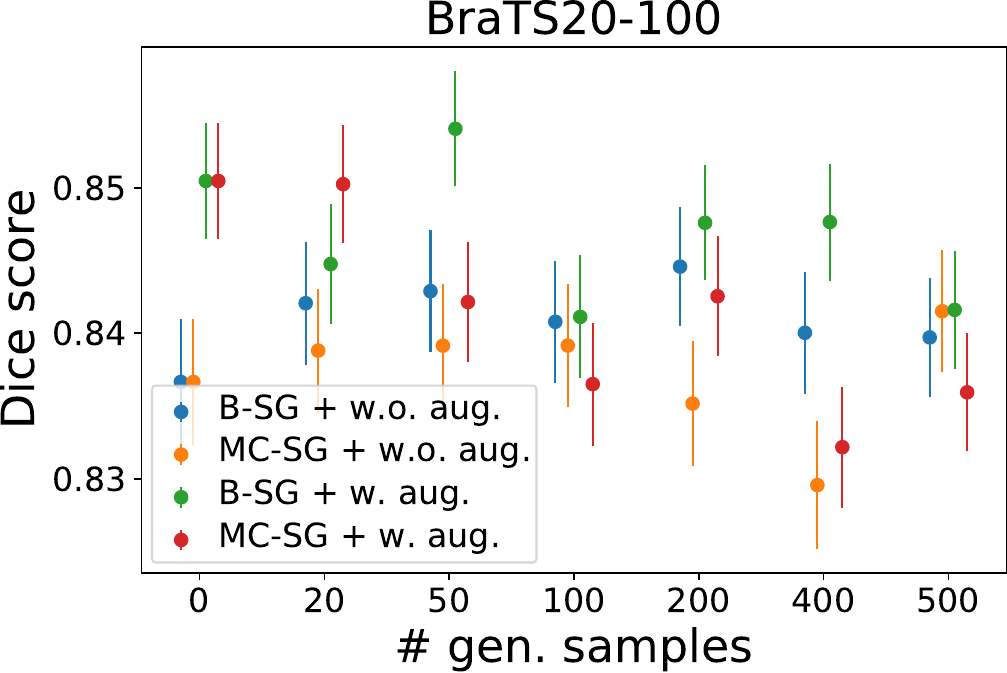}
    \end{subfigure}
    \begin{subfigure}[b]{\factor\textwidth}
		\centering
		\includegraphics[width=\textwidth]{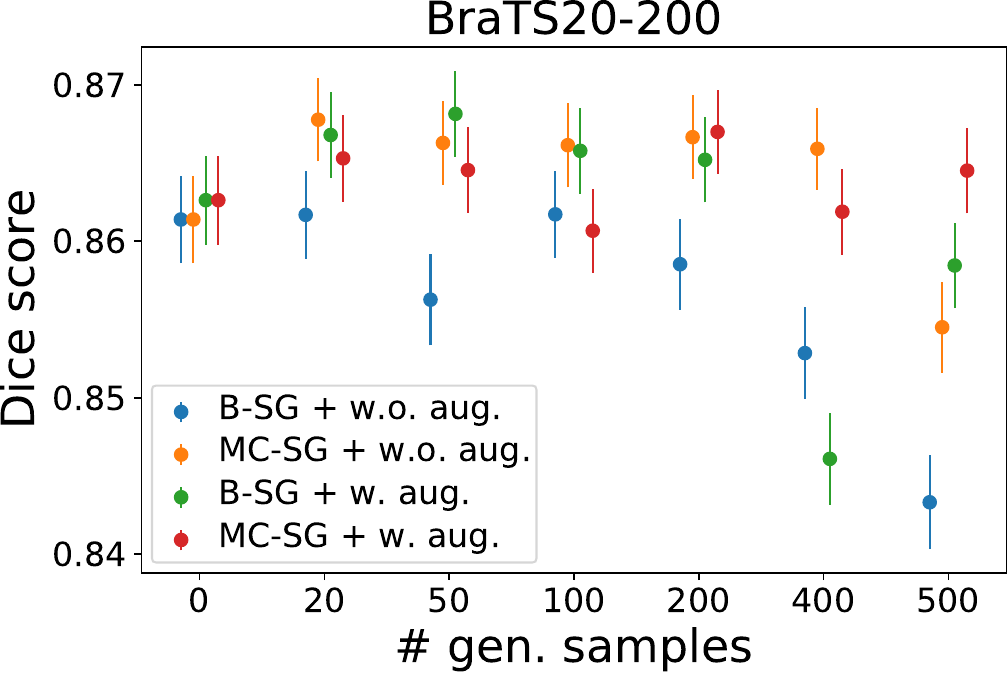}
    \end{subfigure}
    \begin{subfigure}[b]{\factor\textwidth}
		\centering
		\includegraphics[width=\textwidth]{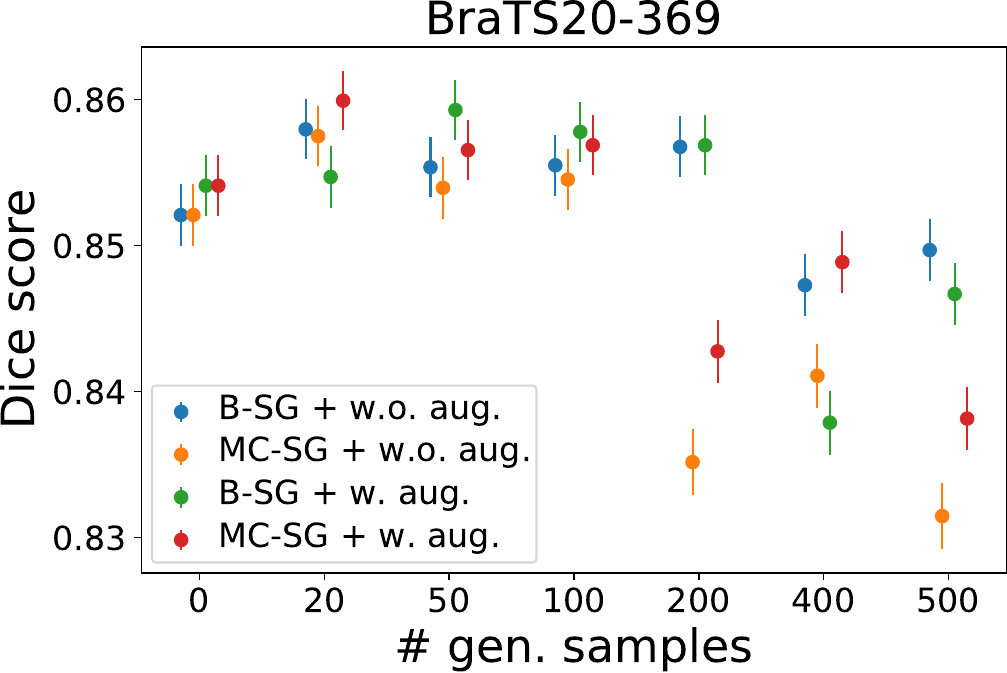}
    \end{subfigure}
    \\
    \vspace{2mm}
    \begin{subfigure}[b]{\factor\textwidth}
		\centering
		\includegraphics[width=\textwidth]{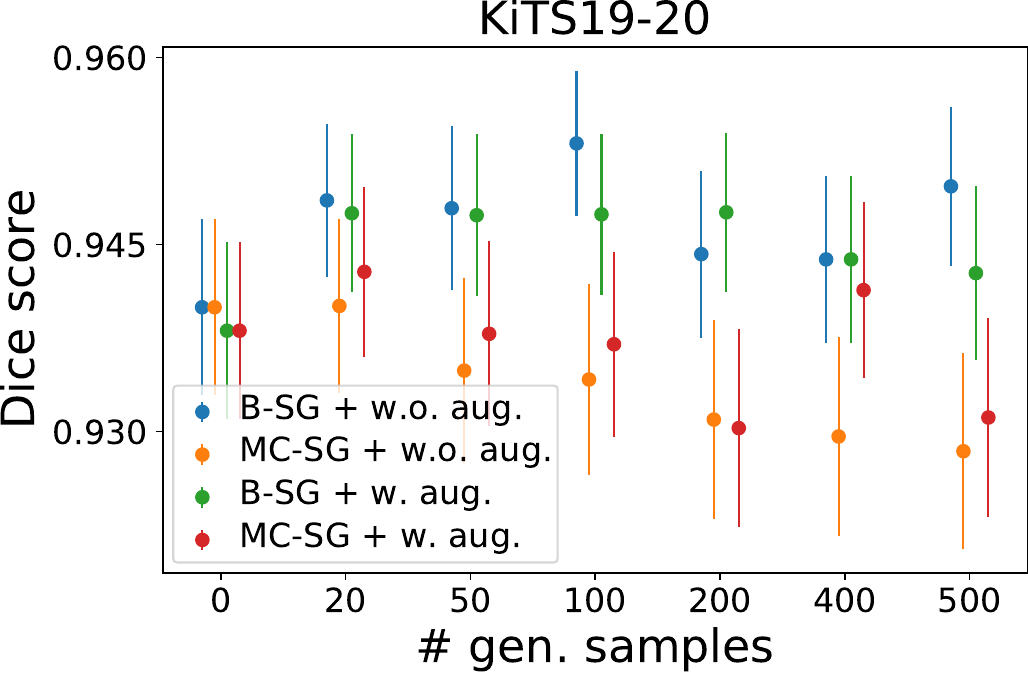}
    \end{subfigure}
    \begin{subfigure}[b]{\factor\textwidth}
		\centering
		\includegraphics[width=\textwidth]{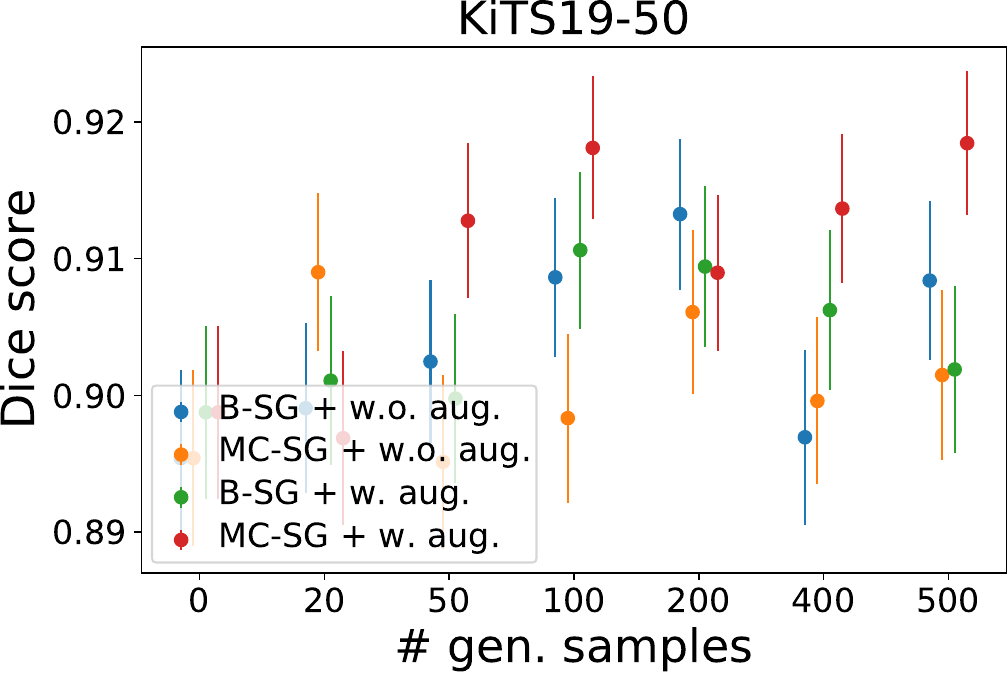}
    \end{subfigure}
    \begin{subfigure}[b]{\factor\textwidth}
		\centering
		\includegraphics[width=\textwidth]{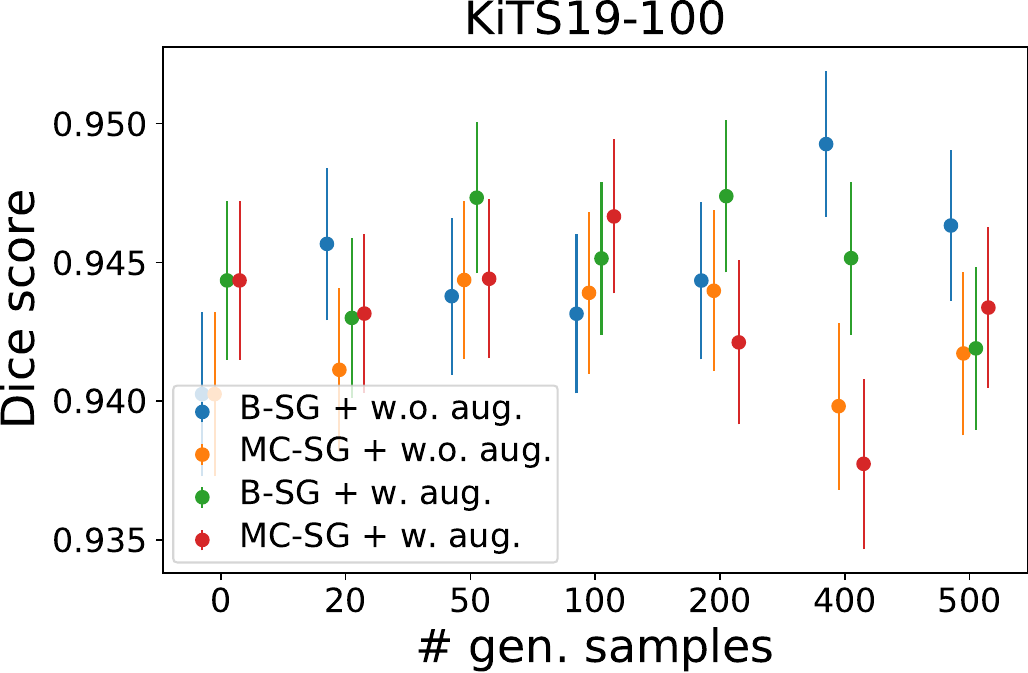}
    \end{subfigure}
    \begin{subfigure}[b]{\factor\textwidth}
		\centering
		\includegraphics[width=\textwidth]{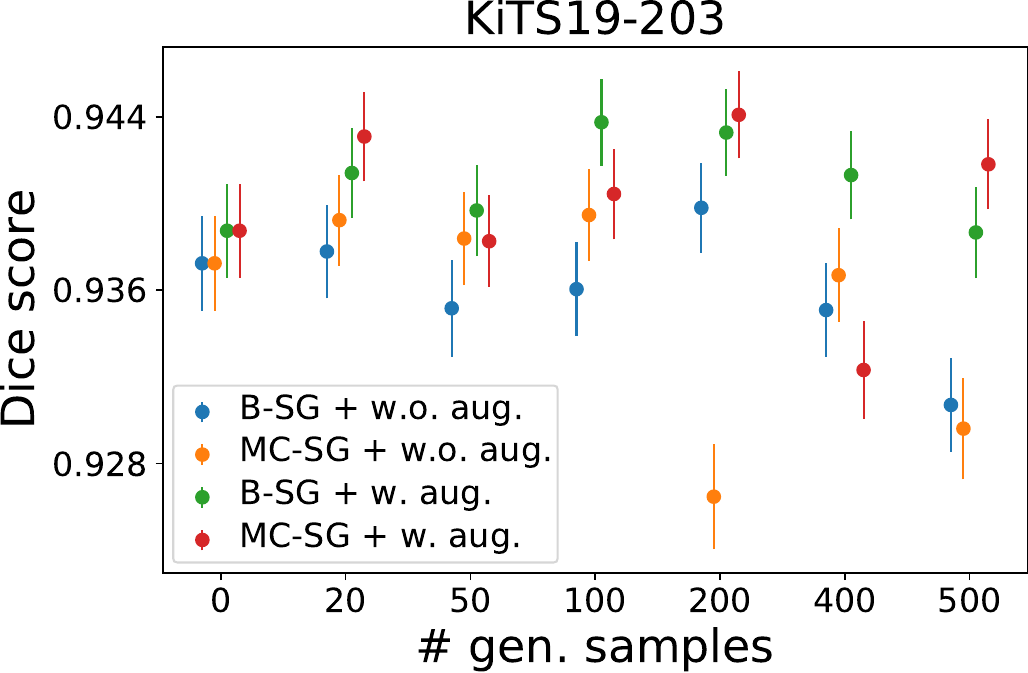}
    \end{subfigure}
    \\
    \vspace{2mm}
    \begin{subfigure}[b]{\factor\textwidth}
		\centering
		\includegraphics[width=\textwidth]{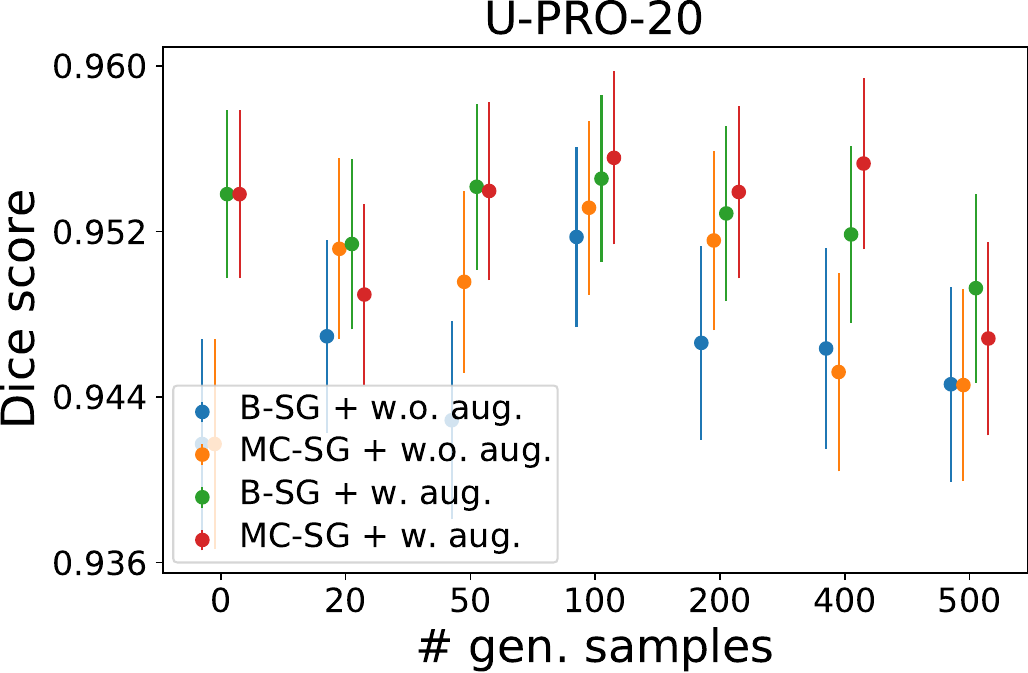}
    \end{subfigure}
    \begin{subfigure}[b]{\factor\textwidth}
		\centering
		\includegraphics[width=\textwidth]{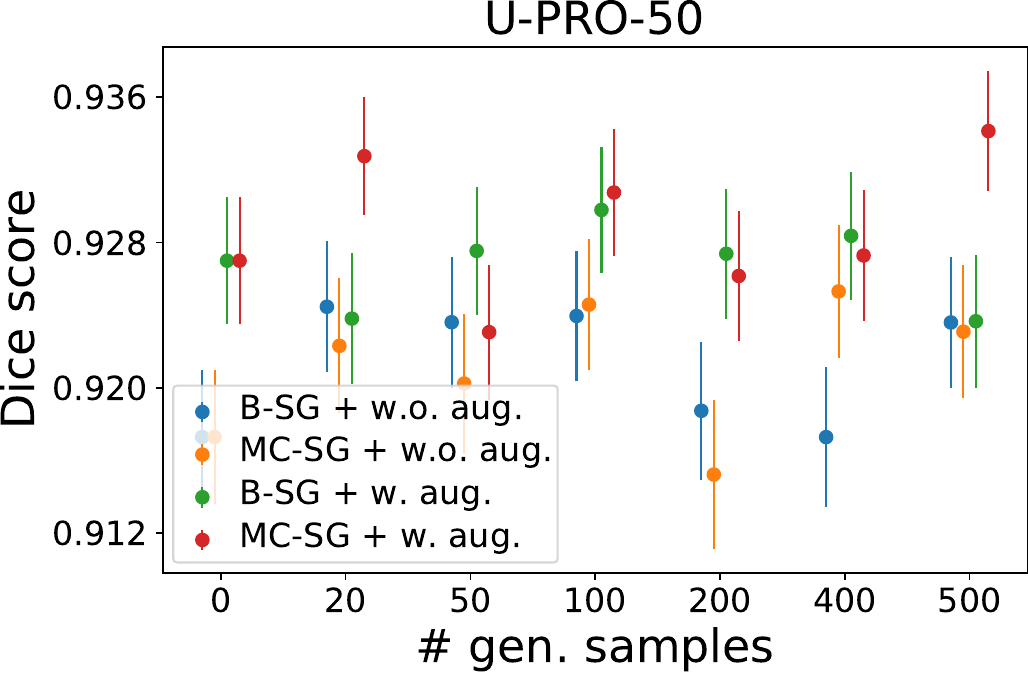}
    \end{subfigure}
    \begin{subfigure}[b]{\factor\textwidth}
		\centering
		\includegraphics[width=\textwidth]{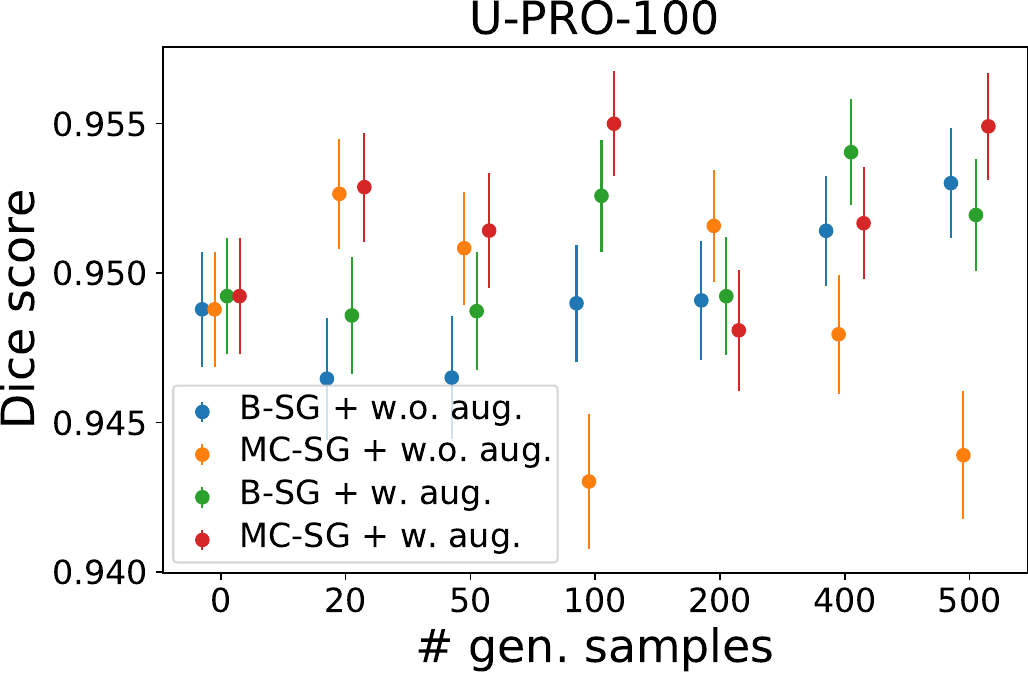}
    \end{subfigure}
    \begin{subfigure}[b]{\factor\textwidth}
		\centering
		\includegraphics[width=\textwidth]{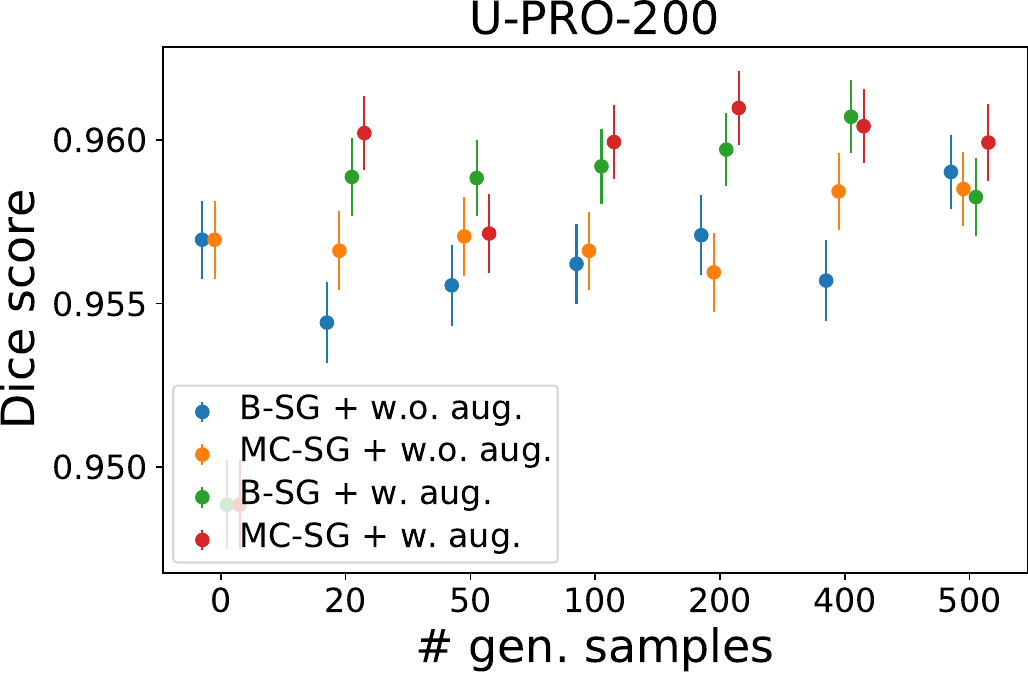}
    \end{subfigure}
    \begin{subfigure}[b]{\factor\textwidth}
		\centering
		\includegraphics[width=\textwidth]{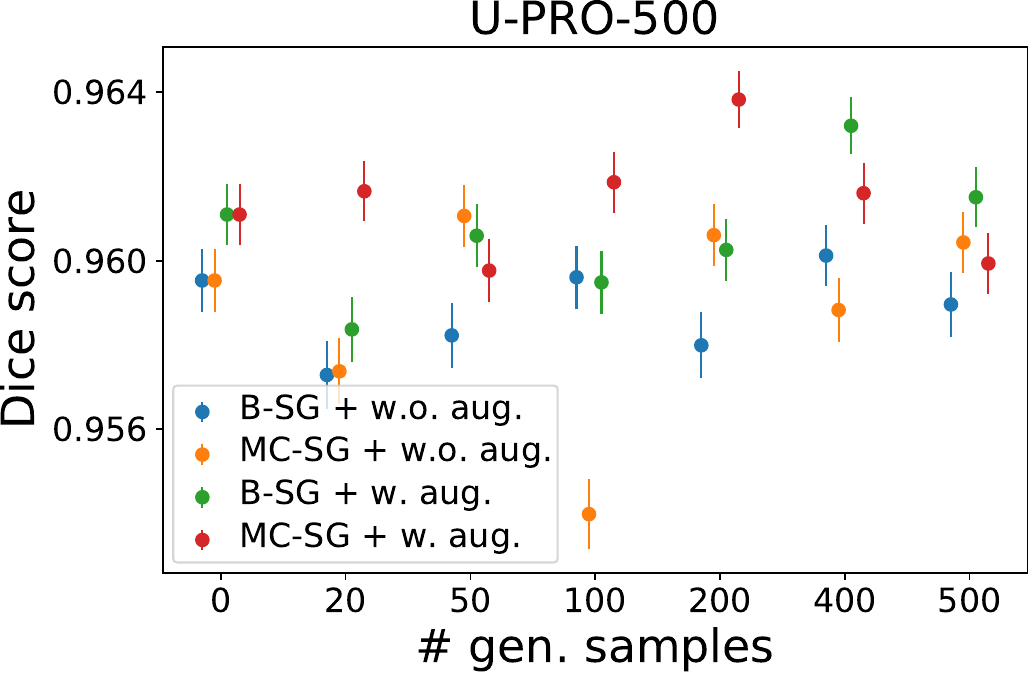}
    \end{subfigure}    
    \caption{Comparison of \gls{dsc} scores and their standard errors of 26 models on 18 datasets (see \tabref{tab:result_nemenyi}) with different numbers of added generated patients. The titles of subfigures present the datasets' names. The $x$-axis of all subfigures shows the number of added generated patients/volumes.}
    \label{fig:gan_result}
\end{figure*}

\Figref{fig:gan_result} presents the \gls{dsc} scores and their standard errors of 26 models on 18 datasets (see \tabref{tab:result_nemenyi}) with different numbers of added generated patients. The titles of subfigures present the datasets' names. For example, \gls{brats}-20 shows that this dataset was created by randomly selecting 20 patients from the \gls{brats} dataset. The $x$-axis of all subfigures indicates the number of added generated patients/volumes. When the ``\# gen. samples'' is equivalent to zero, it represents the baseline (\ie, training without generated samples, see \figref{fig:proposed_pipeline}(a)). The ``w.~aug.'' and  ``w.o.~aug.'' stand for with and without data augmentation---the process of artificially enlarging the size of the training dataset by producing modified versions of the available images.  

\begin{table*}[!th]
    \def\width{1.0cm}
    \centering
	\caption{The results of the Nemenyi post-hoc test comparing all evaluated methods on all datasets. The directions include ($-$), ($+$) and ($0$).  A minus ($-$) means ranked significantly lower. A zero ($0$) means no significant difference. A plus ($+$) means ranked significantly higher when comparing a method to another. The number in the table presents the sum of Nemenyi directions where ($+$) and ($-$) mean adding and subtracting one unit, respectively.}
    \label{tab:result_nemenyi}
    \begin{adjustbox}{max width=.8\textwidth}
    \begin{tabular}{r|rrrrrrrrrrrrrrrrrr}
     Models                              &
    \rotatebox{90}{          U-HEART-20} &
    \rotatebox{90}{         D-SPLEEN-20} &
    \rotatebox{90}{         D-SPLEEN-41} &
    \rotatebox{90}{           IBSR18-18} &
    \rotatebox{90}{           KiTS19-20} &
    \rotatebox{90}{           KiTS19-50} &
    \rotatebox{90}{          KiTS19-100} &
    \rotatebox{90}{          KiTS19-203} &
    \rotatebox{90}{          BraTS20-20} &
    \rotatebox{90}{          BraTS20-50} &
    \rotatebox{90}{         BraTS20-100} &
    \rotatebox{90}{         BraTS20-200} &
    \rotatebox{90}{         BraTS20-369} &
    \rotatebox{90}{            U-PRO-20} &
    \rotatebox{90}{            U-PRO-50} &
    \rotatebox{90}{           U-PRO-100} &
    \rotatebox{90}{           U-PRO-200} &
    \rotatebox{90}{           U-PRO-500} \\
	\cmidrule{1-19}
           BL + w.o. &   0 &   0 &   0 &  10 &   0 &   0 &   0 &   0 &   0 &   0 &   0 &   0 &   1 &   0 &   0 &   0 &   0 &   0  \\
    B-SG + w.o. + 20 &   0 &   0 &   0 &  10 &   0 &   0 &   0 &   0 &   0 &   0 &   0 &   0 &   4 &   0 &   0 &   0 &   0 &   0  \\
    B-SG + w.o. + 50 &   0 &   0 &   0 &  10 &   0 &   0 &   0 &   0 &   0 &   0 &   0 &   0 &   3 &   0 &   0 &   0 &   0 &   0  \\
   B-SG + w.o. + 100 &   0 &   0 &   0 &   6 &   0 &   0 &   0 &   0 &   0 &   0 &   0 &   0 &   3 &   0 &   0 &   0 &   0 &   0  \\
   B-SG + w.o. + 200 &   0 &   0 &   0 &   2 &   0 &   0 &   0 &   0 &   0 &   0 &   0 &   0 &   4 &   0 &   0 &   0 &   0 &   0  \\
   B-SG + w.o. + 400 &   0 &   0 &   0 & -14 &   0 &   0 &   0 &   0 &   0 &   0 &   0 &   0 &   0 &   0 &   0 &   0 &   0 &   0  \\
   B-SG + w.o. + 500 &   0 &   0 &   0 & -20 &   0 &   0 &   0 &   0 &   0 &   0 &   0 &   0 &   0 &   0 &   0 &   0 &   0 &   0  \\
   MC-SG + w.o. + 20 &   0 &   0 &   0 &   9 &   0 &   0 &   0 &   0 &   0 &   0 &   0 &   0 &   4 &   0 &   0 &   0 &   0 &   0  \\
   MC-SG + w.o. + 50 &   0 &   0 &   0 &   6 &   0 &   0 &   0 &   0 &   0 &   0 &   0 &   0 &   1 &   0 &   0 &   0 &   0 &   0  \\
  MC-SG + w.o. + 100 &   0 &   0 &   0 &   2 &   0 &   0 &   0 &   0 &   0 &   0 &   0 &   0 &   1 &   0 &   0 &   0 &   0 &   0  \\
  MC-SG + w.o. + 200 &   0 &   0 &   0 &   3 &   0 &   0 &   0 &   0 &   0 &   0 &   0 &   0 & -12 &   0 &   0 &   0 &   0 &   0  \\
  MC-SG + w.o. + 400 &   0 &   0 &   0 & -16 &   0 &   0 &   0 &   0 &   0 &   0 &   0 &   0 &   0 &   0 &   0 &   0 &   0 &   0  \\
  MC-SG + w.o. + 500 &   0 &   0 &   0 &  -6 &   0 &   0 &   0 &   0 &   0 &   0 &   0 &   0 & -16 &   0 &   0 &   0 &   0 &   0  \\
  \cmidrule{1-19}
             BL + w. &   0 &   0 &   0 &  10 &   0 &   0 &   0 &   0 &   0 &   0 &   0 &   0 &   1 &   0 &   0 &   0 &   0 &   0  \\
      B-SG + w. + 20 &   0 &   0 &   0 &  10 &   0 &   0 &   0 &   0 &   0 &   0 &   0 &   0 &   3 &   0 &   0 &   0 &   0 &   0  \\
      B-SG + w. + 50 &   0 &   0 &   0 &   8 &   0 &   0 &   0 &   0 &   0 &   0 &   0 &   0 &   4 &   0 &   0 &   0 &   0 &   0  \\
     B-SG + w. + 100 &   0 &   0 &   0 &   5 &   0 &   0 &   0 &   0 &   0 &   0 &   0 &   0 &   4 &   0 &   0 &   0 &   0 &   0  \\
     B-SG + w. + 200 &   0 &   0 &   0 & -12 &   0 &   0 &   0 &   0 &   0 &   0 &   0 &   0 &   4 &   0 &   0 &   0 &   0 &   0  \\
     B-SG + w. + 400 &   0 &   0 &   0 &  -6 &   0 &   0 &   0 &   0 &   0 &   0 &   0 &   0 &  -7 &   0 &   0 &   0 &   0 &   0  \\
     B-SG + w. + 500 &   0 &   0 &   0 & -12 &   0 &   0 &   0 &   0 &   0 &   0 &   0 &   0 &   0 &   0 &   0 &   0 &   0 &   0  \\
     MC-SG + w. + 20 &   0 &   0 &   0 &   6 &   0 &   0 &   0 &   0 &   0 &   0 &   0 &   0 &   4 &   0 &   0 &   0 &   0 &   0  \\
     MC-SG + w. + 50 &   0 &   0 &   0 &   5 &   0 &   0 &   0 &   0 &   0 &   0 &   0 &   0 &   3 &   0 &   0 &   0 &   0 &   0  \\
    MC-SG + w. + 100 &   0 &   0 &   0 &   3 &   0 &   0 &   0 &   0 &   0 &   0 &   0 &   0 &   3 &   0 &   0 &   0 &   0 &   0  \\
    MC-SG + w. + 200 &   0 &   0 &   0 &  -5 &   0 &   0 &   0 &   0 &   0 &   0 &   0 &   0 &   0 &   0 &   0 &   0 &   0 &   0  \\
    MC-SG + w. + 400 &   0 &   0 &   0 & -10 &   0 &   0 &   0 &   0 &   0 &   0 &   0 &   0 &   0 &   0 &   0 &   0 &   0 &   0  \\
    MC-SG + w. + 500 &   0 &   0 &   0 &  -4 &   0 &   0 &   0 &   0 &   0 &   0 &   0 &   0 & -12 &   0 &   0 &   0 &   0 &   0  \\
    \bottomrule
    \end{tabular}
    \end{adjustbox}
\end{table*}

\tabref{tab:result_nemenyi} shows the Nemenyi post-hoc test comparing all evaluated methods (row) on all datasets (column). The number in the table presents the sum of Nemenyi directions where ($+$) and ($-$) mean adding and subtracting one unit, respectively. The Nemenyi directions include ($-$), ($+$) and ($0$).  The minus ($-$) means ranked significantly lower. The zero ($0$) means no significant difference. The plus ($+$) means ranked significantly higher when comparing a method to another method (see \eg, \tabref{tab:result_nemenyi_brats369}). From  \tabref{tab:result_nemenyi}, we can determine how well a method compares to other evaluated methods. 

From \tabref{tab:result_nemenyi}, we can see that there are only two datasets that we can determine the significant differences from the methods, which are IBSR18-18 and BraTS20-369. For the other datasets, the differences between evaluated methods are non-significant (all zeros), \ie, the segmentation networks did not benefit from the generated images by the proposed \glspl{gan}. Looking at the IBSR18-18 and BraTS20-369 in \tabref{tab:result_nemenyi}, it is apparent that the baselines with and without data augmentation are among the top-performing methods. Another observation is that adding many generated images did not improve the performance; otherwise, it was even worse in some cases, \eg, ``B-SG  w.o.~+ 500''.

\begin{table*}[!th]
    \def\width{1.0cm}
    \centering
    \caption{The results of the Nemenyi post-hoc test comparing all evaluated methods on the \gls{brats}-369. A minus ($-$) means ranked significantly lower, a zero ($0$) means non-significant difference, and a plus ($+$) means ranked significantly higher, when comparing a method in the rows to a method in the columns. ``BL'' is the baseline (training without generated images). ``B-SG'' is the \sg2. ``MC-SG'' is the proposed method. The ``w.o.'' denotes without augmentation, while the ``w.'' denotes with augmentation. The number following after ``+ w.'' or ``+ w.'' is the number of generated patients.}
    \label{tab:result_nemenyi_brats369}    
    \begin{adjustbox}{max width=\textwidth}
    \begin{tabular}{rccccccccccccccccccccccccccr}
	&
	\rotatebox{90}{           BL + w.o.} &
	\rotatebox{90}{    B-SG + w.o. + 20} &
	\rotatebox{90}{    B-SG + w.o. + 50} &
	\rotatebox{90}{   B-SG + w.o. + 100} &
	\rotatebox{90}{   B-SG + w.o. + 200} &
	\rotatebox{90}{   B-SG + w.o. + 400} &
	\rotatebox{90}{   B-SG + w.o. + 500} &
	\rotatebox{90}{   MC-SG + w.o. + 20} &
	\rotatebox{90}{   MC-SG + w.o. + 50} &
	\rotatebox{90}{  MC-SG + w.o. + 100} &
	\rotatebox{90}{  MC-SG + w.o. + 200} &
	\rotatebox{90}{  MC-SG + w.o. + 400} &
	\rotatebox{90}{  MC-SG + w.o. + 500} &
	\rotatebox{90}{             BL + w.} &
	\rotatebox{90}{      B-SG + w. + 20} &
	\rotatebox{90}{      B-SG + w. + 50} &
	\rotatebox{90}{     B-SG + w. + 100} &
	\rotatebox{90}{     B-SG + w. + 200} &
	\rotatebox{90}{     B-SG + w. + 400} &
	\rotatebox{90}{     B-SG + w. + 500} &
	\rotatebox{90}{     MC-SG + w. + 20} &
	\rotatebox{90}{     MC-SG + w. + 50} &
	\rotatebox{90}{    MC-SG + w. + 100} &
	\rotatebox{90}{    MC-SG + w. + 200} &
	\rotatebox{90}{    MC-SG + w. + 400} &
	\rotatebox{90}{    MC-SG + w. + 500} &
	\rotatebox{90}{               Score} \\
	\cmidrule{1-28}
           BL + w.o. & ~~~ & $0$ & $0$ & $0$ & $0$ & $0$ & $0$ & $0$ & $0$ & $0$ & $0$ & $0$ & $+$ & $0$ & $0$ & $0$ & $0$ & $0$ & $0$ & $0$ & $0$ & $0$ & $0$ & $0$ & $0$ & $0$ &   1  \\
    B-SG + w.o. + 20 & $0$ & ~~~ & $0$ & $0$ & $0$ & $0$ & $0$ & $0$ & $0$ & $0$ & $+$ & $0$ & $+$ & $0$ & $0$ & $0$ & $0$ & $0$ & $+$ & $0$ & $0$ & $0$ & $0$ & $0$ & $0$ & $+$ &   4  \\
    B-SG + w.o. + 50 & $0$ & $0$ & ~~~ & $0$ & $0$ & $0$ & $0$ & $0$ & $0$ & $0$ & $+$ & $0$ & $+$ & $0$ & $0$ & $0$ & $0$ & $0$ & $0$ & $0$ & $0$ & $0$ & $0$ & $0$ & $0$ & $+$ &   3  \\
   B-SG + w.o. + 100 & $0$ & $0$ & $0$ & ~~~ & $0$ & $0$ & $0$ & $0$ & $0$ & $0$ & $+$ & $0$ & $+$ & $0$ & $0$ & $0$ & $0$ & $0$ & $0$ & $0$ & $0$ & $0$ & $0$ & $0$ & $0$ & $+$ &   3  \\
   B-SG + w.o. + 200 & $0$ & $0$ & $0$ & $0$ & ~~~ & $0$ & $0$ & $0$ & $0$ & $0$ & $+$ & $0$ & $+$ & $0$ & $0$ & $0$ & $0$ & $0$ & $+$ & $0$ & $0$ & $0$ & $0$ & $0$ & $0$ & $+$ &   4  \\
   B-SG + w.o. + 400 & $0$ & $0$ & $0$ & $0$ & $0$ & ~~~ & $0$ & $0$ & $0$ & $0$ & $0$ & $0$ & $0$ & $0$ & $0$ & $0$ & $0$ & $0$ & $0$ & $0$ & $0$ & $0$ & $0$ & $0$ & $0$ & $0$ &   0  \\
   B-SG + w.o. + 500 & $0$ & $0$ & $0$ & $0$ & $0$ & $0$ & ~~~ & $0$ & $0$ & $0$ & $0$ & $0$ & $0$ & $0$ & $0$ & $0$ & $0$ & $0$ & $0$ & $0$ & $0$ & $0$ & $0$ & $0$ & $0$ & $0$ &   0  \\
   MC-SG + w.o. + 20 & $0$ & $0$ & $0$ & $0$ & $0$ & $0$ & $0$ & ~~~ & $0$ & $0$ & $+$ & $0$ & $+$ & $0$ & $0$ & $0$ & $0$ & $0$ & $+$ & $0$ & $0$ & $0$ & $0$ & $0$ & $0$ & $+$ &   4  \\
   MC-SG + w.o. + 50 & $0$ & $0$ & $0$ & $0$ & $0$ & $0$ & $0$ & $0$ & ~~~ & $0$ & $0$ & $0$ & $+$ & $0$ & $0$ & $0$ & $0$ & $0$ & $0$ & $0$ & $0$ & $0$ & $0$ & $0$ & $0$ & $0$ &   1  \\
  MC-SG + w.o. + 100 & $0$ & $0$ & $0$ & $0$ & $0$ & $0$ & $0$ & $0$ & $0$ & ~~~ & $0$ & $0$ & $+$ & $0$ & $0$ & $0$ & $0$ & $0$ & $0$ & $0$ & $0$ & $0$ & $0$ & $0$ & $0$ & $0$ &   1  \\
  MC-SG + w.o. + 200 & $0$ & $-$ & $-$ & $-$ & $-$ & $0$ & $0$ & $-$ & $0$ & $0$ & ~~~ & $0$ & $0$ & $0$ & $-$ & $-$ & $-$ & $-$ & $0$ & $0$ & $-$ & $-$ & $-$ & $0$ & $0$ & $0$ & -12  \\
  MC-SG + w.o. + 400 & $0$ & $0$ & $0$ & $0$ & $0$ & $0$ & $0$ & $0$ & $0$ & $0$ & $0$ & ~~~ & $0$ & $0$ & $0$ & $0$ & $0$ & $0$ & $0$ & $0$ & $0$ & $0$ & $0$ & $0$ & $0$ & $0$ &   0  \\
  MC-SG + w.o. + 500 & $-$ & $-$ & $-$ & $-$ & $-$ & $0$ & $0$ & $-$ & $-$ & $-$ & $0$ & $0$ & ~~~ & $-$ & $-$ & $-$ & $-$ & $-$ & $0$ & $0$ & $-$ & $-$ & $-$ & $0$ & $0$ & $0$ & -16  \\
             BL + w. & $0$ & $0$ & $0$ & $0$ & $0$ & $0$ & $0$ & $0$ & $0$ & $0$ & $0$ & $0$ & $+$ & ~~~ & $0$ & $0$ & $0$ & $0$ & $0$ & $0$ & $0$ & $0$ & $0$ & $0$ & $0$ & $0$ &   1  \\
      B-SG + w. + 20 & $0$ & $0$ & $0$ & $0$ & $0$ & $0$ & $0$ & $0$ & $0$ & $0$ & $+$ & $0$ & $+$ & $0$ & ~~~ & $0$ & $0$ & $0$ & $0$ & $0$ & $0$ & $0$ & $0$ & $0$ & $0$ & $+$ &   3  \\
      B-SG + w. + 50 & $0$ & $0$ & $0$ & $0$ & $0$ & $0$ & $0$ & $0$ & $0$ & $0$ & $+$ & $0$ & $+$ & $0$ & $0$ & ~~~ & $0$ & $0$ & $+$ & $0$ & $0$ & $0$ & $0$ & $0$ & $0$ & $+$ &   4  \\
     B-SG + w. + 100 & $0$ & $0$ & $0$ & $0$ & $0$ & $0$ & $0$ & $0$ & $0$ & $0$ & $+$ & $0$ & $+$ & $0$ & $0$ & $0$ & ~~~ & $0$ & $+$ & $0$ & $0$ & $0$ & $0$ & $0$ & $0$ & $+$ &   4  \\
     B-SG + w. + 200 & $0$ & $0$ & $0$ & $0$ & $0$ & $0$ & $0$ & $0$ & $0$ & $0$ & $+$ & $0$ & $+$ & $0$ & $0$ & $0$ & $0$ & ~~~ & $+$ & $0$ & $0$ & $0$ & $0$ & $0$ & $0$ & $+$ &   4  \\
     B-SG + w. + 400 & $0$ & $-$ & $0$ & $0$ & $-$ & $0$ & $0$ & $-$ & $0$ & $0$ & $0$ & $0$ & $0$ & $0$ & $0$ & $-$ & $-$ & $-$ & ~~~ & $0$ & $-$ & $0$ & $0$ & $0$ & $0$ & $0$ &  -7  \\
     B-SG + w. + 500 & $0$ & $0$ & $0$ & $0$ & $0$ & $0$ & $0$ & $0$ & $0$ & $0$ & $0$ & $0$ & $0$ & $0$ & $0$ & $0$ & $0$ & $0$ & $0$ & ~~~ & $0$ & $0$ & $0$ & $0$ & $0$ & $0$ &   0  \\
     MC-SG + w. + 20 & $0$ & $0$ & $0$ & $0$ & $0$ & $0$ & $0$ & $0$ & $0$ & $0$ & $+$ & $0$ & $+$ & $0$ & $0$ & $0$ & $0$ & $0$ & $+$ & $0$ & ~~~ & $0$ & $0$ & $0$ & $0$ & $+$ &   4  \\
     MC-SG + w. + 50 & $0$ & $0$ & $0$ & $0$ & $0$ & $0$ & $0$ & $0$ & $0$ & $0$ & $+$ & $0$ & $+$ & $0$ & $0$ & $0$ & $0$ & $0$ & $0$ & $0$ & $0$ & ~~~ & $0$ & $0$ & $0$ & $+$ &   3  \\
    MC-SG + w. + 100 & $0$ & $0$ & $0$ & $0$ & $0$ & $0$ & $0$ & $0$ & $0$ & $0$ & $+$ & $0$ & $+$ & $0$ & $0$ & $0$ & $0$ & $0$ & $0$ & $0$ & $0$ & $0$ & ~~~ & $0$ & $0$ & $+$ &   3  \\
    MC-SG + w. + 200 & $0$ & $0$ & $0$ & $0$ & $0$ & $0$ & $0$ & $0$ & $0$ & $0$ & $0$ & $0$ & $0$ & $0$ & $0$ & $0$ & $0$ & $0$ & $0$ & $0$ & $0$ & $0$ & $0$ & ~~~ & $0$ & $0$ &   0  \\
    MC-SG + w. + 400 & $0$ & $0$ & $0$ & $0$ & $0$ & $0$ & $0$ & $0$ & $0$ & $0$ & $0$ & $0$ & $0$ & $0$ & $0$ & $0$ & $0$ & $0$ & $0$ & $0$ & $0$ & $0$ & $0$ & $0$ & ~~~ & $0$ &   0  \\
    MC-SG + w. + 500 & $0$ & $-$ & $-$ & $-$ & $-$ & $0$ & $0$ & $-$ & $0$ & $0$ & $0$ & $0$ & $0$ & $0$ & $-$ & $-$ & $-$ & $-$ & $0$ & $0$ & $-$ & $-$ & $-$ & $0$ & $0$ & ~~~ & -12  \\
    \bottomrule
    \end{tabular}
    \end{adjustbox}
\end{table*}

\section{Conclusion}

In this work, we introduced a novel conditional \acrlong{gan} based on the standard StyleGAN2 to generate high-quality medical images with different modalities. We evaluated the quality of the generated medical images and the effect of this augmentation on the segmentation performance. There are, in fact, more investigations and experiments needed for conclusions on the effect of generated images on the downstream segmentation task.

\bibliographystyle{plainnat}
\bibliography{refs}



\end{document}